\documentclass[runningheads]{llncs}

\usepackage[mobile]{accv}

\usepackage{accvabbrv}
\usepackage{microtype}
\usepackage{comment}
\usepackage{subcaption}
\usepackage{booktabs} 
\usepackage{placeins}
\usepackage{longtable}
\usepackage{multirow}
\usepackage{array,makecell}
\usepackage{siunitx} 
\usepackage{graphicx}
\usepackage{amsmath}
\usepackage{amssymb}

\usepackage[accsupp]{axessibility}  

\usepackage[pagebackref,breaklinks,colorlinks,citecolor=accvblue]{hyperref}

\usepackage{orcidlink}

\begin{document}

\title{Smooth Deep Saliency} 

\titlerunning{Smooth Deep Saliency}

\author{Rudolf Herdt \and
Maximilian Schmidt \and
Daniel Otero Baguer \and
Peter Maass}

\authorrunning{Herdt et al.}

\institute{University of Bremen
\email{rherdt@uni-bremen.de}}

\maketitle

\begin{abstract}
In this work, we investigate methods to reduce the noise in deep saliency maps coming from convolutional downsampling.
Those methods make the investigated models more interpretable for gradient-based saliency maps, computed in hidden layers.
We evaluate the faithfulness of those methods using insertion and deletion metrics, finding that saliency maps computed in hidden layers perform better compared to both the input layer and GradCAM.
We test our approach on different models trained for image classification on ImageNet1K, and models trained for tumor detection on Camelyon16 and in-house real-world digital pathology scans of stained tissue samples.
Our results show that the checkerboard noise in the gradient gets reduced, resulting in smoother and therefore easier to interpret saliency maps.\footnote{The code is available at https://github.com/rherdt185/smooth-deep-saliency}
\end{abstract}

\section{Introduction}

For critical applications, such as in the healthcare sector, it is important to gain an insight into how the model arrives at its decision.
To gain such an understanding, saliency methods can be used, ideally highlighting the structures on which the model relies to support its prediction.
Saliency methods are usually applied at either the input or the last convolutional layer \cite{Selvaraju_2017_ICCV}.
Their application in the input layer leads to noisy results, and in the final convolution layer, the saliency map lacks detail.
In this paper, we focus on the middle ground and apply saliency methods in hidden layers.
This gives us smoother results while preserving the details. As a result, we can use high-quality saliency maps to identify individual cell nuclei, which are used by trained models to detect tumors.
On the one hand, using hidden layers also allows us to observe the transition from lower-level features, like the outline of cell nuclei in a Resnet18 \cite{resnets} trained on Camelyon16 \cite{Ehteshami_Bejnordi2017-dk} data or individual cell nuclei for a segmentation model trained on digital pathology (Digipath) data at a lower magnification, highlighted in earlier layers. On the other hand, it also gives access to higher-level features in deeper layers, e.g. segmenting the whole cell nuclei for the Resnet18 model or structures formed by cells for the model trained on Digipath data.

A problem in computing gradient-based saliency maps in hidden layers of a model that uses convolutional downsampling is the checkerboard noise this downsampling introduces to its gradient, similar to what transposed convolutions exhibit in the forward pass \cite{odena2016deconvolution}.
This noise dominates the resulting saliency map and makes them very difficult to interpret.
Therefore, in this paper, we investigate several methods to remove that checkerboard noise, which allows us to use saliency methods in hidden layers and test them on several models and datasets.

With a stride 2 convolution (downsampling), not every input pixel gets multiplied by every weight.
Due to the stride of 2, every other pixel is skipped, which can introduce a checkerboard pattern with a periodicity of 2 in the gradient (see \cref{sec:conv-noise}).
%
%
Since that noise is coming from the convolutional downsampling, we can mitigate it by changing the backward pass of each convolutional downsampling layer, in order to let each input pixel receive a gradient by each weight.
The other method we use is to train a bilinear surrogate model, where we replace each convolutional downsampling with a bilinear downsampling combined with two trainable stride 1 convolutions, one before and after the bilinear downsampling. These are then trained to match the behavior of the original model.
That way, we can compare the accuracy and predictions of the surrogate model to the original model. 
%
%
%
The output of each surrogate path should match the output of the convolutional downsampling it replaces under the L1 loss.
The bilinear surrogate model is trained on the training data of the original model.
We show that the accuracy and predictions of the surrogate model closely match those of the original model, while the generated saliency maps have less noise, as measured by total variation.
And we test the saliency maps by insertion and deletion metrics introduced in \cite{petsiuk2018rise}, and observe that the bilinear surrogate model still has comparable insertion and deletion metrics compared to the original model.

%

%
%
%
%

%

%
%
%


\section{Related Work}

%
%
%

Saliency maps are used to highlight structures in the input that the model utilized for its prediction.
%
%
%
%
%
The saliency methods could be divided into three groups.
First, those that only utilize activations like CAM \cite{Oquab_2015_CVPR}, secondly those that also utilize gradients of backpropagation like Grad-CAM \cite{Selvaraju_2017_ICCV}, DeepLift \cite{shrikumar2019learning}, Integrated Gradients \cite{pmlr-v70-sundararajan17a}, or simply using the gradient \cite{simonyan2014deep}.
And lastly pertubation base methods like RISE \cite{petsiuk2018rise}, which are usually applied at the input.
Pertubation-based methods generally require several evaluations of the model, which makes them slow.
To get gradient-based methods more stable SmoothGrad \cite{SmoothGrad} can be employed, which runs the saliency method several times each time adding Gaussian noise to the input and then returning the average saliency map.
%

Gradient-based saliency methods are usually applied at either the input layer or the last convolutional layer, like \cite{Selvaraju_2017_ICCV}, whereas we focus on computing the saliency maps from hidden layers of the model.
\cite{olah2018the} also computes saliency maps at hidden layers and couples them with feature visualization.
To mitigate noise in the saliency maps, they employ Gaussian blurring.
\cite{Rao_2022_CVPR} also used Gaussian blurring to smooth saliency maps of hidden layers.
We limit ourselves to models that utilize convolutional downsampling and, instead of using Gaussian blurring, we work on removing the source of the noise.

Saliency methods can be invariant of what the model has learned, which is undesired \cite{SanityCheckSaliencyMaps}.
In order to measure the faithfulness of the saliency maps generated with our methods, we use deletion and insertion metrics as introduced in \cite{petsiuk2018rise}.
Using our methods, we also observe a drop in the metrics, if the model weights are randomized, suggesting the methods are sensitive to what the model has learned.
\cite{olah2017feature} visualized neurons of a CNN, and observed that the visualizations become more high-level the deeper the neuron is located in the model.
We observe something similar in the attribution for the Camelyon16 and Digipath models, where in earlier layers the cell nuclei are outlined and in deeper layers segmented (Camelyon16).
And the saliency map for the Digipath model highlights individual cell nuclei in earlier layers and structures formed by them in deeper layers.

\subsection{Checkerboard Noise from Stride 2 Convolution}
\label{sec:conv-noise}

\begin{figure}[htb]
\begin{center}
\centerline{\includegraphics[width=0.15\textwidth]{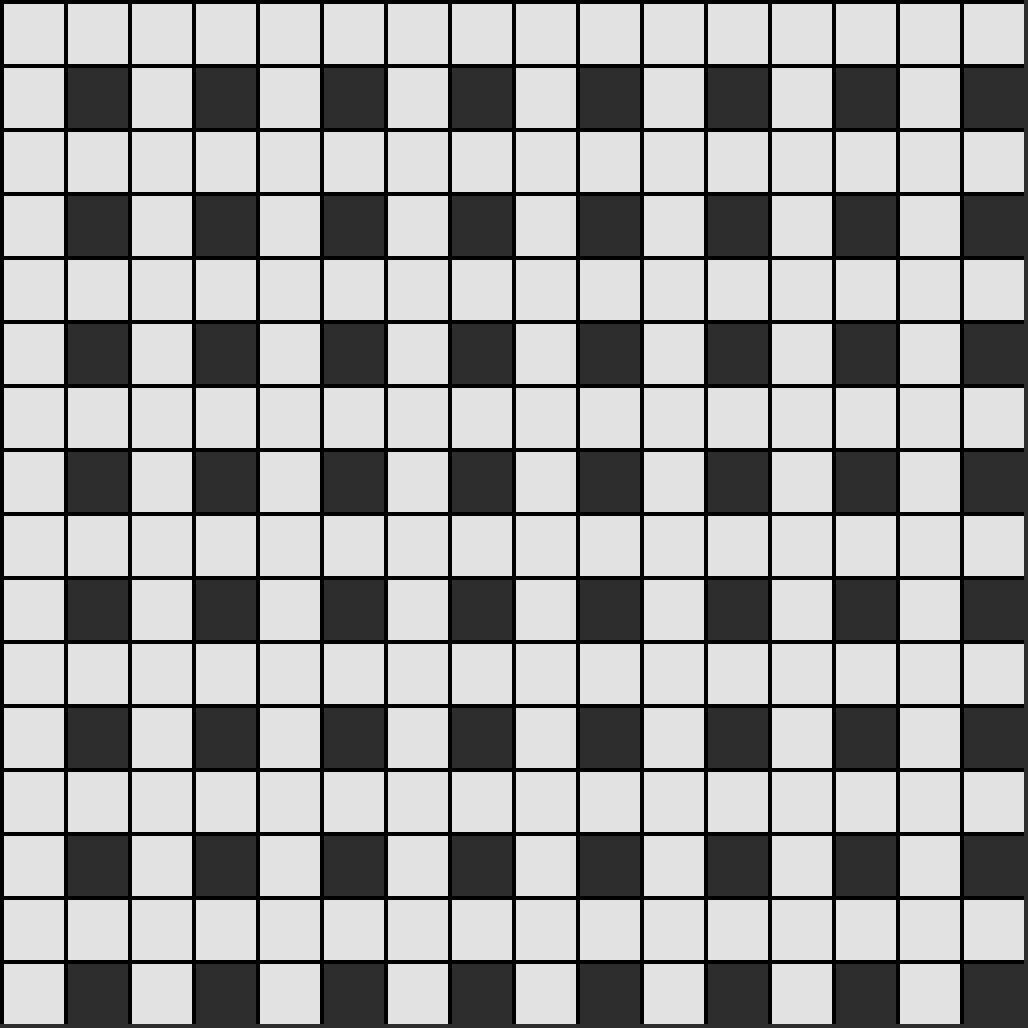}}
\caption{Checkerboard pattern in the gradient of the $16\times16$ image.}
\label{fig:checkerboard_grad}
\end{center}
\vskip -0.2in
\end{figure}

With a stride 2 convolution, not every pixel is multiplied by every weight.
Due to the stride of 2, every other pixel is skipped.
%
%
As an example, let $x$ be a matrix of ones with a size of $16\times16$, $k = \begin{pmatrix} 1 & 1\\ 1 & -1 \end{pmatrix}$ and $*$ the operator for a stride 2 convolution.
\cref{fig:checkerboard_grad} shows the result of calculating $\nabla_x (\sum x * k)$, i.e.\ the gradient towards $x$ when computing the sum of the convolution of $x$ with kernel $k$.
We can see a periodic checkerboard pattern in the gradient.
The black pixels were multiplied by $-1.0$ from the kernel and, therefore, received a gradient of -1.0. In contrast, the white pixels were multiplied by the $1.0$ value from the kernel and received a gradient of 1.0.
That is the noise pattern we mitigate with our methods.

\section{Methods}
\label{sec:methods}

In this section, we show an overview of our bilinear surrogate method.
In addition, we describe the evaluation metrics we use and how we compute the saliency maps.

\subsection{Overview}

\begin{figure*}[htb]
\begin{center}
\centerline{\includegraphics[width=0.7\textwidth]{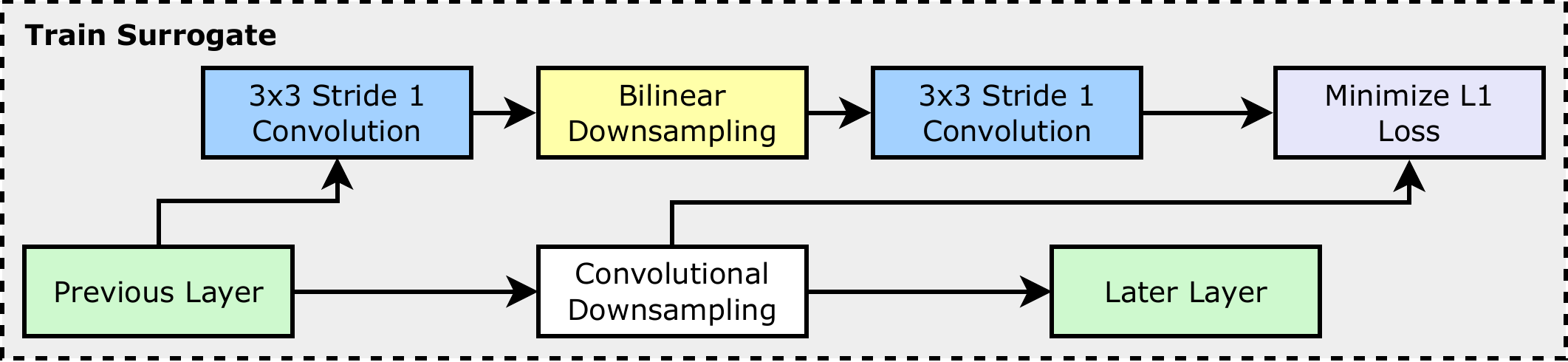}}
\caption{Training the bilinear surrogate. The weights of the two $3\times3$ stride 1 convolutions are not shared, they are different convolutions.}
\label{fig:train_surrogate}
\end{center}
\vskip -0.2in
\end{figure*}

To reduce the checkerboard noise in the gradient coming from the convolutional downsampling, we investigate three different methods:

\begin{itemize}
    \item \textbf{Bilinear Surrogate:} Replacing each convolutional downsampling with a bilinear surrogate path (consisting of two stride 1 convolutions with a bilinear downsampling in between them), as shown in \cref{fig:train_surrogate}.
    This approach requires training data, in order to train the surrogate paths.

    \item \textbf{Backward Hook:} Changing the backward pass of each convolutional downsampling, by taking the gradients it would normally propagate backward, rolling them 4 times spatially, and returning the mean of those 4 tensors.
    We roll them by (h, w) pixels with h and w being (0, 0), (0, 1), (1, 0), (1, 1).
    The rolling means that the line of pixels that would be shifted outside of the image, moves to the opposite side of the tensor.

    \item \textbf{Forward Hook:} Changing the forward pass of each convolutional downsampling in a way that we get the gradients of the second method by default.
    This means we run the forward pass 4 times, each time with a spatially rolled input, and return the mean of those 4 tensors.
    We roll them the same as in the method mentioned directly above, namely by (0, 0), (0, 1), (1, 0), (1, 1) pixels.
    %
    %
    Although changing only the backward pass as described in the method above is enough to remove the checkerboard noise, changing the forward pass allows us to compare the accuracy and predictions against the original model.

\end{itemize}

\begin{figure}[htb]
\begin{center}
\centerline{\includegraphics[width=0.5\textwidth]{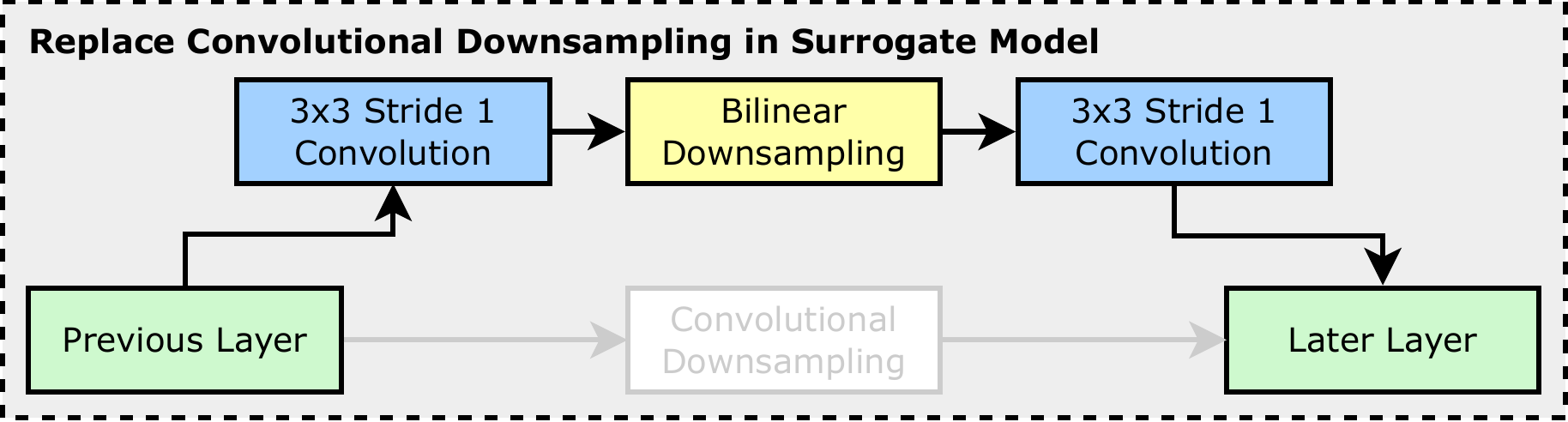}}
\caption{Using the bilinear surrogate path in evaluation. The convolutional downsamplings will be ignored, they are replaced by the bilinear surrogate.}
\label{fig:use_surrogate}
\end{center}
\vskip -0.2in
\end{figure}

%
%
%

%
With the bilinear surrogate approach, all convolutional downsamplings except the first one in the model are replaced.
Similarly, with the forward and backward hook approach, we do not change the behavior of the first convolutional downsampling.
We do not replace the first one, since we always compute the saliency maps towards a deeper layer, therefore we do not need to replace it.

In training of the bilinear surrogate model, we replace each convolutional downsampling individually (i.e. we only use the convolutional downsampling in the forward pass and not the bilinear surrogate one).
But in the evaluation, we use them all together, as visible from \cref{fig:use_surrogate}.
We use the L1 loss as a loss function in training.

\subsection{Evaluation}

With our bilinear surrogate method, we aim to achieve three goals: 

\begin{itemize}
    \item Reduce the checkerboard noise in the gradient to make saliency maps in hidden layers more interpretable.
    \item Match the accuracy of the original model. Because the idea would be to use the bilinear surrogate model instead of the original model. Therefore, the accuracy needs to be as good as the original model.
    \item Have faithful saliency maps. If the saliency maps for the bilinear surrogate model loose faithfulness compared to the original model they would be less useful.
\end{itemize}

To validate whether we succeeded in reducing the checkerboard noise in the gradient, we can visually inspect some saliency maps.
Further, we also run total variation (TV) on the saliency maps over a validation dataset.
To evaluate how well we match the prediction behavior of the original model, we compare the accuracy and/or difference in prediction for only the target class and all classes of the original and the surrogate model on a validation dataset.
For the difference in prediction, we use the output of the model after softmax.
Then we compute the difference in prediction twice: Once only for the target class and once for all classes.
To measure the faithfulness of the saliency methods we use insertion and deletion metrics introduced in \cite{petsiuk2018rise}.

\subsection{Computing the Saliency Map}

We use and compare four methods to compute the saliency maps, namely DeepLift, Integrated Gradients (IG), simply computing the gradient and GradCAM.
We use the first three methods at both the input layer and at hidden layers, whereas we use GradCAM only at the last convolutional layer.
As baseline we always use a black image (zeros), after normalization (which would correspond to a gray image for the ImageNet models before normalization).
The saliency map is computed using the logits output of the model (before softmax or sigmoid).

For image classification, we compute the saliency map against the ground truth label.
For semantic segmentation, if we only have the background class as a label, we compute the saliency map against the background label. Otherwise, we compute it against the majority non-background label.

\subsection{Total Variation}


In this section, we describe how we compute the total variation (TV) to measure the noise in the saliency maps.
To evaluate how much noise our methods remove, we compute the TV of the saliency map from the original model and the surrogate model and compare the values.
Since we only want to measure the noise and not the scale of it, we first process the saliency map before measuring TV.
Let $z$ be the saliency map, then for each channel $z_c$ separately, we first set the mean to zero,

\begin{equation}
  z'_c = z_c - \mu(z_c)
  \label{eq:important}
\end{equation}

and then divide by the mean of the absolute value and compute the mean over the channels (with $n$ being the number of channels)

\begin{equation}
  \sum_c \frac{1}{n} \text{TV}\big( \frac{z_c'}{\text{max}(\mu(\vert z_c' \vert), \epsilon)} \big)
  \label{eq:tv}
\end{equation}

$\epsilon$ is set to $10^{-6}$ and is used to avoid division by zero (i.e. if $z_c'$ is zero everywhere).
\cref{eq:tv} is the resulting total variation reported in the experiments section.

\section{Experiments}
\label{sec:experiments}

\subsection{Setup}

In this section we describe the hardware, datasets and models we use in our experiments.

\subsubsection{Hardware and Software}

For all our experiments, we use a Linux server with 8 Nvidia RTX 2080 Ti GPUs.
On the software side, we use PyTorch \cite{paszke2019} version 1.13.1 with CUDA version 11.6 and torchvision version 0.14.1.
For computing the saliency maps we use the Captum library \cite{kokhlikyan2020captum} version 0.6.0.

\subsubsection{Training}

%
The forward and backward hook approaches work on the original model, and require no additional training.
All networks in our experiments are running in eval mode. We only train the bilinear surrogate paths, each consisting of two 3x3 stride 1 convolutions, with a bilinear downsampling in between them. The rest of the network is kept frozen.
%
%
We train the bilinear surrogate for 10 epochs.
As optimizer we use Adam \cite{kingma_adam_2015} with a learning rate of 0.001.
As batch size we use 64 for training the surrogate models on the ImageNet1K \cite{deng_imagenet_2009} data, 512 for the Camelyon16 data and 16 for the Digipath data (due to the larger patch size of 1536 we use a smaller batch size here).

\begin{table}[tb]
    \caption{Train- and validation set sizes for the three datasets ImageNet1K, Camelyon16, and Digipath.}
    \label{tab:datasets}
    \begin{center}
    \begin{small}
    \begin{sc}
    \vskip -0.2in
    \resizebox{.5\textwidth}{!}{
    \begin{tabular}{l|cccc}  
    \toprule
     & ImageNet1K & Camelyon16 & Digipath\\
    \midrule
    Training Samples   & \num{1281167} & \num{9655756} & \num{129634} \\
    Validation Samples & \num{50000} & \num{46267899} & \num{899} \\  
    TV Validation Samples & \num{500} & \num{500} & \num{899} \\  
    Used Image Size &  \num{224}x\num{224} &  \num{224}x\num{224} &  \num{1536}x\num{1536} \\ 
    Training Epochs &  \num{10} &  \num{10} &  \num{10} \\ 
    \bottomrule
    \end{tabular}
    }
    \end{sc}
    \end{small}
    \end{center}
    \vskip -0.1in
\end{table}

\begin{table}[tb]
\caption{Accuracy on the ImageNet1K validation data.}
\label{tab:imagenet-accuracy}
\begin{center}
\begin{small}
\begin{sc}
\vskip -0.2in
\resizebox{0.5\textwidth}{!}{
\begin{tabular}{l|c|c|c}  
\toprule
Network & \makecell{Original \\ Accuracy} & \makecell{Surrogate \\ Accuracy} & \makecell{Forward Hook \\ Accuracy} \\
\midrule
 ResNet34    & 73.29 & 73.12 & 53.25 \\
 ResNet50    & 76.15 & 76.05 & 73.00 \\
 ResNet50 Robust    & 57.89 & 57.86 & 56.10 \\
\bottomrule
\end{tabular}
}
\end{sc}
\end{small}
\end{center}
\end{table}

\begin{table*}[tb]
\caption{Prediction difference between original and bilinear surrogate model, measured after softmax.}
\label{tab:class-prediction}
\begin{center}
\begin{small}
\begin{sc}
\vskip -0.2in
\resizebox{0.75\textwidth}{!}{
\begin{tabular}{l|c|c|c|c}  
\toprule
 & \multicolumn{2}{c}{\textbf{Bilinear Surrogate}} & \multicolumn{2}{c}{\textbf{Forward Hook}} \\ 
Network & All Classes & Only Target Class & All Classes & Only Target Class \\
\midrule
 ResNet34    & \textbf{9.52} $\pm$ 12.31 & \textbf{2.39} $\pm$ 4.08 & 95.84 $\pm$ 66.84 & 30.26 $\pm$ 31.33 \\
 ResNet50    & \textbf{8.90} $\pm$ 12.48 & \textbf{2.38} $\pm$ 4.25 & 39.44 $\pm$ 45.58 & 11.38 $\pm$ 17.30 \\
 ResNet50 Robust    & \textbf{5.84} $\pm$ 4.37 & \textbf{0.97} $\pm$ 1.39 & 38.55 $\pm$ 25.62 & 7.95 $\pm$ 10.52 \\
 ResNet18 Camelyon16    & \textbf{0.41} $\pm$ 1.40 & \textbf{0.41} $\pm$ 1.40 & 3.63 $\pm$ 10.04 & 3.63 $\pm$ 10.04 \\
 ResNet34 Digipath      & \textbf{4.05} $\pm$ 2.11 & \textbf{2.04} $\pm$ 1.09 & 37.33 $\pm$ 34.36 & 19.76 $\pm$ 16.15 \\
\bottomrule
\end{tabular}
}
\end{sc}
\end{small}
\end{center}
\vskip -0.1in
\end{table*}

\subsubsection{Datasets}

We evaluate our method on three datasets:

\begin{itemize}
    \item \textbf{ImageNet1K}.
    We evaluate the accuracy and the difference in prediction between the original and surrogate model on the \num{50000} validation images. The total variation and saliency metrics are evaluated on every hundredth image of the validation data, resulting in 500 images.
    
    \item \textbf{Camelyon16}.
    For training we use all the patches of the original training data were the original model predicts tumor (score after sigmoid {\textgreater} 0), plus randomly sample the same amount of patches were the original model is not predicting tumor (score after sigmoid {\textless} 0).
    The difference in prediction is evaluated on all the \num{46267899} validation samples. The total variation and saliency metrics are evaluated on 500 samples were the original model predicts tumor randomly chosen out of the validation samples.

    \item \textbf{Digipath}. This is an in-house dataset which is not openly available.
    Here the training data consists of \num{129634} images.
    We evaluate the difference in prediction and total variation on 899 images not used in training of the original or surrogate model.
\end{itemize}

\subsubsection{Models}

On the ImageNet1K data we use a ResNet34 and a ResNet50 both openly available from the pytorch model zoo.
Further, we use an adversarially robust ResNet50 from \cite{NEURIPS2019_6f2268bd}.
For the Camelyon16 data we use an openly available ResNet18 from the monai model zoo.
And for the Digipath data we use an in-house U-Net segmentation model with a ResNet34 backbone.
While introduced a while ago, the ResNet50 architecture is still utilized in recent work \cite{Prashanth_2024_WACV}.


%
%

\subsection{Accuracy and Prediction Difference}

\cref{tab:imagenet-accuracy} shows the accuracy of the original, bilinear surrogate and forward hooked models, on the ImageNet1K data.
For the bilinear surrogate models we observe a maximum decrease of accuracy of 0.17 for the ResNet34 model, and a minimum decrease of 0.03 for the robust ResNet50 model.
The forward hooked models have a larger decrease of accuracy, especially in the case of the ResNet34 model where the accuracy decreases by 20.04.

In \cref{tab:class-prediction} we evaluate how the prediction of the models changes, once for the sum over all predicted classes and once only for the target class.
We report the differences in softmax values times 100 (i.e. a reported difference of 9.52 for the ResNet34 model would mean a difference of 0.0952).
For the bilinear surrogate models we observe the maximum change of prediction for the sum of all classes and only for the target class of 9.52 respectively 2.39 for the Resnet34 model.
That aligns with the drop of accuracy being largest for the ResNet34 surrogate model.
Averaged over all models, the forward hooked models have a 7.8 times higher difference in prediction for all classes, and a 8.8 times higher difference in prediction for the target class compared to the bilinear surrogate model.

\begin{table}[tb]
    \caption{Best insertion and deletion layers and methods.}
    \label{tab:insertion_deletion}
    \begin{center}
    \begin{small}
    \begin{sc}
    \vskip -0.2in
    \resizebox{.75\textwidth}{!}{
    \begin{tabular}{l|l|cccc}  
    \toprule
     Network & Score & DeepLift & IG & Grad & GradCAM \\
    \midrule
    \multirow{2}{*}{ResNet34}
     & Ins  $\uparrow$ & \textbf{0.679} (S)(2\_3) & $0.663$ (S)(2\_3) & $0.559$ (F)(4\_0) & $0.648$ \\
     & Del $\downarrow$ & \textbf{0.045} (O)(1\_2) & $0.052$ (O)(1\_2) & $0.081$ (O)(1\_2) & $0.127$ \\
     \midrule
    \multirow{2}{*}{ResNet50}
     & Ins & \textbf{0.706} (S)(3\_1) & $0.683$ (S)(4\_2) & $0.531$ (B)(3\_4) & $0.684$ \\
     & Del & \textbf{0.044} (O)(1\_2) & $0.052$ (O)(1\_2) & $0.071$ (O)(1\_1) & $0.147$ \\
    \midrule
    \multirow{2}{*}{\makecell{ResNet50 \\ Robust}}
     & Ins & $0.492$ (O)(Input) & \textbf{0.544} (O)(Input) & $0.430$ (O)(Input) & $0.397$ \\
     & Del & \textbf{0.041} (O)(1\_1) & $0.042$ (O)(Input) & $0.070$ (O)(3\_2) & $0.066$ \\
    \midrule
    \multirow{2}{*}{\makecell{ResNet18 \\ Camelyon}}
     & Ins & $0.738$ (S)(1\_1) & \textbf{0.775} (O)(1\_0) & 0.630 (O)(1\_2) & $0.625$ \\
     & Del & \textbf{0.022} (O)(Input) & $0.030$ (O)(Input) & 0.062 (O)(2\_0) & $0.366$ \\
     
    \bottomrule
    \end{tabular}
    }
    \end{sc}
    \end{small}
    \end{center}
    \vskip -0.1in
\end{table}

\begin{figure*}[htbp]
    \centering
    \begin{tabular}{ccc}
        \begin{subfigure}{0.33\textwidth}
            \centering
            \includegraphics[width=\textwidth]{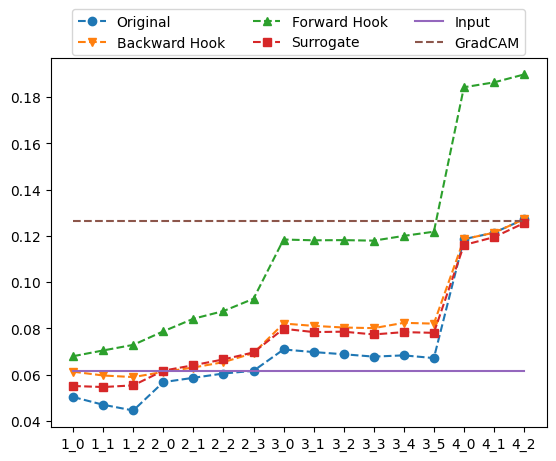}
            \caption{ResNet34 del $\downarrow$}
        \end{subfigure} &
        \begin{subfigure}{0.33\textwidth}
            \centering
            \includegraphics[width=\textwidth]{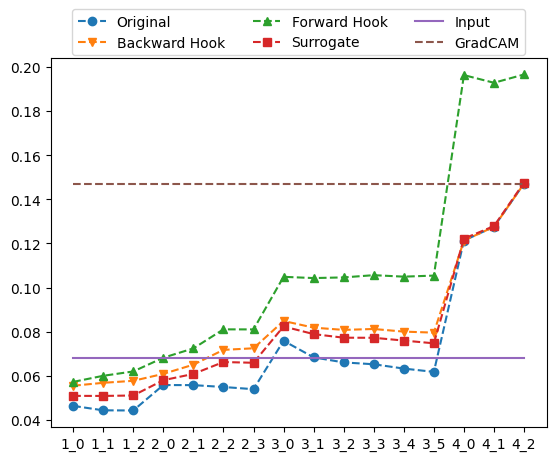}
            \caption{ResNet50 del $\downarrow$}
        \end{subfigure} &
        \begin{subfigure}{0.33\textwidth}
            \centering
            \includegraphics[width=\textwidth]{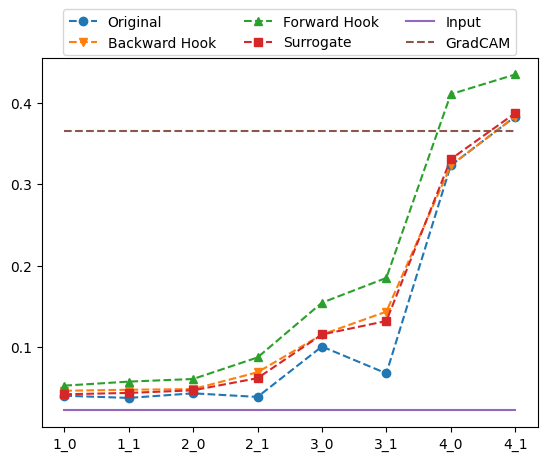}
            \caption{ResNet18 del $\downarrow$}
        \end{subfigure} \\
        
        \begin{subfigure}{0.33\textwidth}
            \centering
            \includegraphics[width=\textwidth]{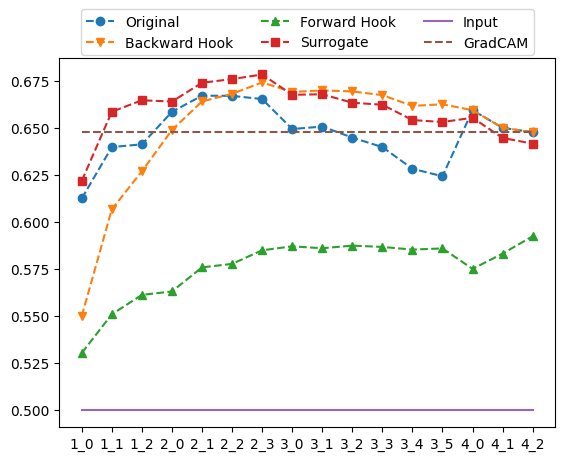}
            \caption{ResNet34 ins $\uparrow$}
        \end{subfigure} &
        \begin{subfigure}{0.33\textwidth}
            \centering
            \includegraphics[width=\textwidth]{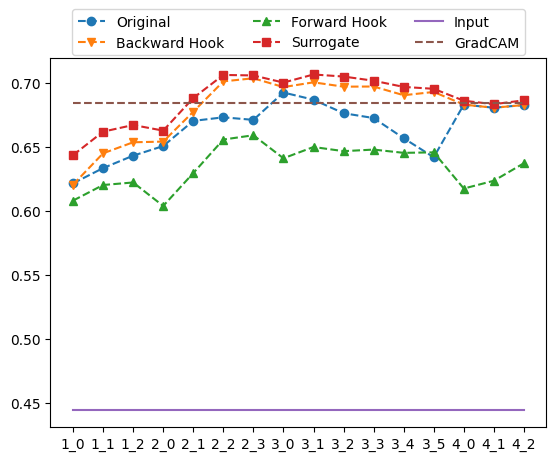}
            \caption{ResNet50 ins $\uparrow$}
        \end{subfigure} &
        \begin{subfigure}{0.33\textwidth}
            \centering
            \includegraphics[width=\textwidth]{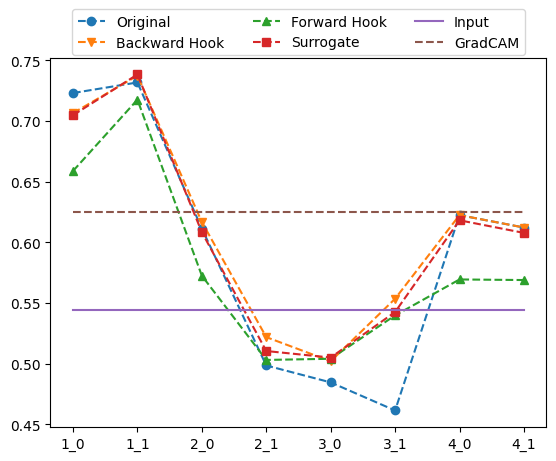}
            \caption{ResNet18 ins $\uparrow$}
        \end{subfigure}
    \end{tabular}
    \caption{Insertion and Deletion metrics using DeepLift.}
    \label{fig:insertion_deletion}
\end{figure*}

\subsection{Insertion and Deletion}

In this section we evaluate the saliency maps based on the insertion and deletion metrics proposed in \cite{petsiuk2018rise} and use their implementation.
%
%
That is, for deletion we set the deleted pixels to gray ((0, 0, 0) after normalization) for the ImageNet and Camelyon16 models (same value that we used as baseline in the attribution methods).
For insertion we use a blurred version of the original image as starting point.

\cref{fig:insertion_deletion} shows the insertion and deletion scores for DeepLift for the ResNet34, ResNet50 and ResNet18 Camelyon16 models.
On the x-axis is the layer, going deeper into the model from left to right.
As baselines we use DeepLift computed towards the input layer, and GradCAM computed at the last convolutional layer (layer 4\_2) both times using the original model (results for the other saliency methods as well as for the robust ResNet50 can be found in the supplementary material).

\cref{tab:insertion_deletion} shows the layer and method that performs best for insertion and deletion for each of the four saliency methods.
Each entry can be read as: Metric result (method)(layer).
The method can be one of four: Original model (O), bilinear surrogate (S), backward hooked (B), forward hooked (F).
For GradCAM we always use the original model and the last convolutional layer.
In total either DeepLift or Integrated Gradient performs best, with DeepLift performing best 6 out of 8 times.
Further attribution in hidden layers performs better than in the input layer (6 out of 8 times).
Of those 6 times where a hidden layer performs best, twice the bilinear surrogate model performs best (both times in insertion).

Averaged over the ImageNet1K models and layers, the attribution with DeepLift in hidden layers for the bilinear surrogate model has a 2.14\% higher insertion score compared to the original model.
But it also has an 11.8\% higher deletion score compared to the original model.
The reason for the worse deletion score could likely be that in the original model some pixels count more than others due to the convolutional downsampling.
Therefore the model may be relying on less input pixels for making its predictions, whereas the bilinear surrogate relies on the pixels more evenly.
The backward hook method has an 0.84\% higher insertion score, but a 16.2\% higher deletion score.
And the forward hook method has both a 7.02\% lower insertion score, and a 44.2\% higher deletion score, making it considerably less faithful compared to using the original model.

\subsection{Randomize Model Weights}

The result of saliency methods should depend on what the model has learned, since we want to explain the trained model.
\cite{SanityCheckSaliencyMaps} found that some saliency methods, are invariant of what the model has learned and rather depend on the architecture, those saliency methods give similar results for trained and randomly initialized models.
Therefore, we test our methods with the model randomization test.
We set the layers of the trained model to the random initialized version, starting from the end of the model (from the fully connected layer).

\begin{figure}[tb]
    \centering
    \begin{tabular}{ll}
        \begin{subfigure}{0.3\textwidth}
            \centering
            \includegraphics[width=\textwidth]{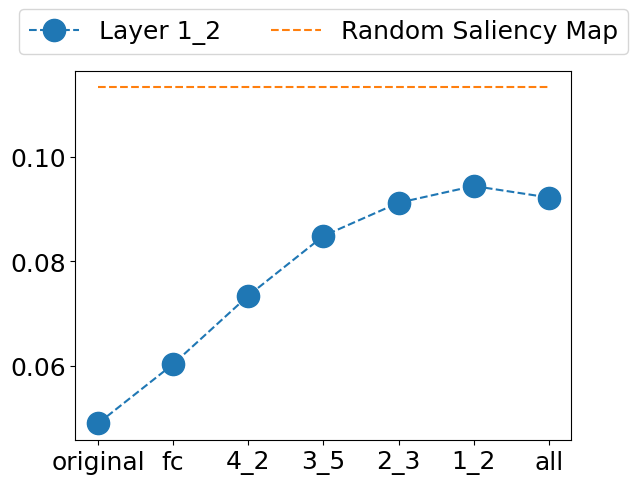}
            \caption*{Deletion Layer1}
        \end{subfigure} &
        \begin{subfigure}{0.3\textwidth}
            \centering
            \includegraphics[width=\textwidth]{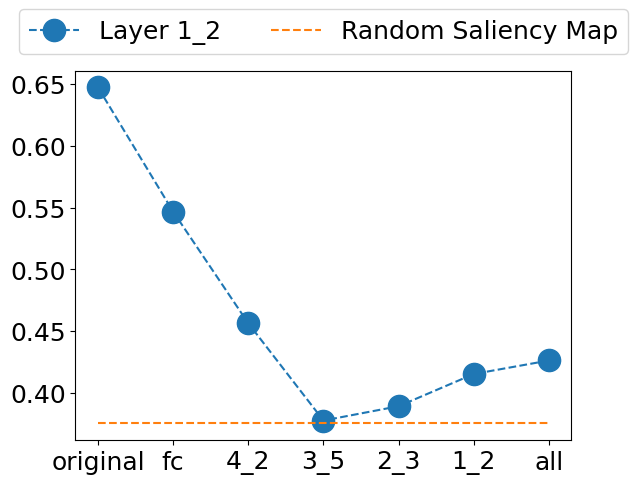}
            \caption*{Insertion Layer1}
        \end{subfigure}
    \end{tabular}
    \caption{Insertion and Deletion scores for randomizing the model, using DeepLift as attribution method.}
    \label{fig:model_randomization}
\end{figure}

\cref{fig:model_randomization} shows the results for the ResNet34 model on ImageNet1K data.
The figure shows the insertion and deletion scores when computing the attribution at Layer1 using DeepLift.
The bottom of each sub-figure shows till which layer the model is randomized.
The metrics generally become worse, the more layers are randomized.
For the deletion metrics, using a completely randomized model still performs a bit better compared to using a random saliency map (gaussian noise at the spatial size of the layer, upscaled to the input size), suggesting the model architecture has some impact on the result.
But overall the metrics become considerably worse when the model is randomized, which we see as confirmation that the method is sensitive to what the model has learned.

\begin{figure*}[htb]
    \centering
    \begin{subfigure}{0.32\textwidth}
        \centering
        \includegraphics[width=\textwidth]{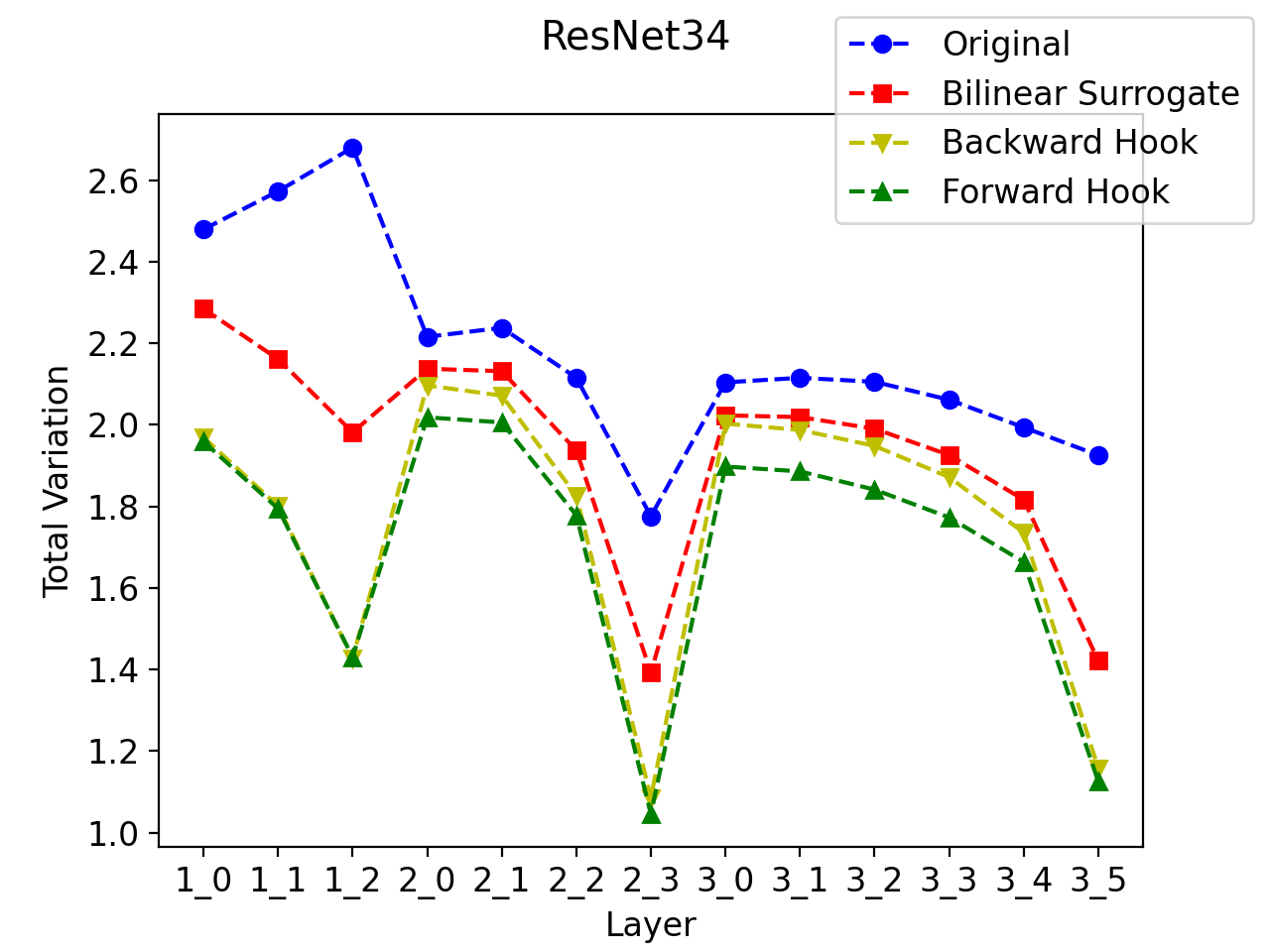}
    \end{subfigure}
    \begin{subfigure}{0.32\textwidth}
        \centering
        \includegraphics[width=\textwidth]{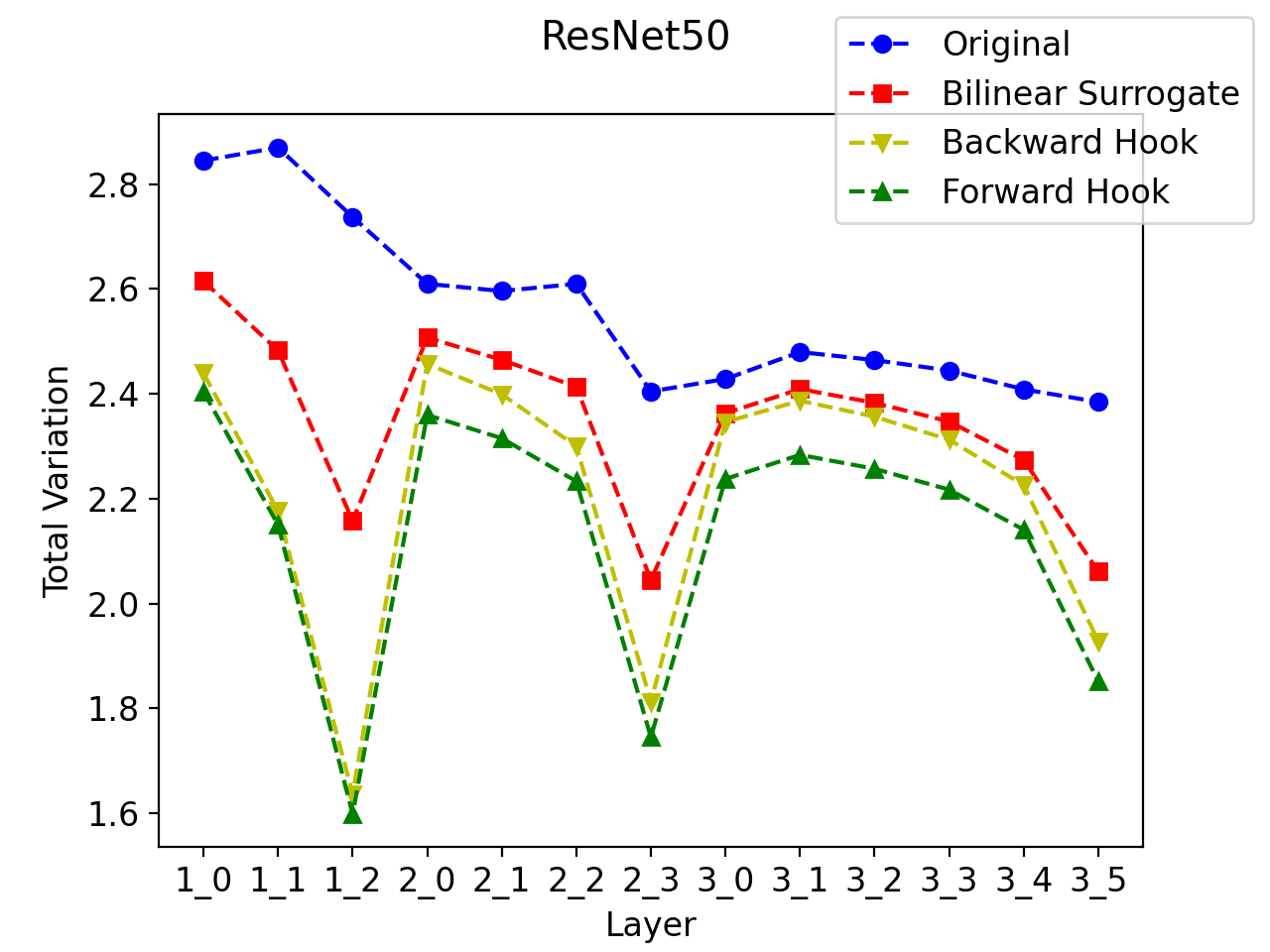}
    \end{subfigure}
    \begin{subfigure}{0.32\textwidth}
        \centering
        \includegraphics[width=\textwidth]{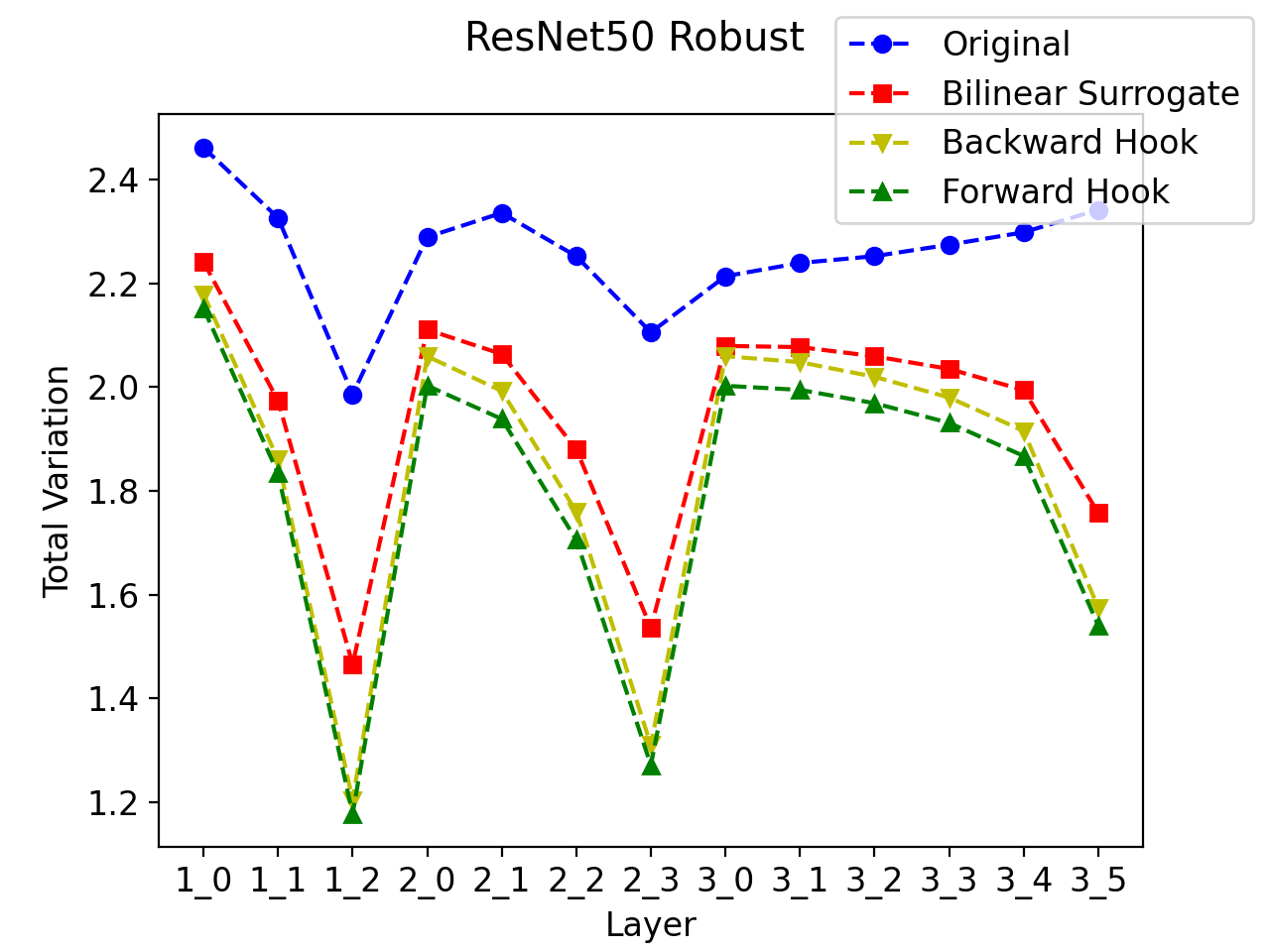}
    \end{subfigure}
    \caption{Total variation of the gradient saliency map method.}
    \label{fig:TV_imagenet_layers}
\end{figure*}

\begin{table}[tb]
    \caption{Percentage reduction in total variation of the saliency map compared to the original model.}
    \label{tab:tv_aggregate}
    \begin{center}
    \begin{small}
    \begin{sc}
    \vskip -0.2in
    \resizebox{.5\textwidth}{!}{
    \begin{tabular}{l|l|cccc}  
    \toprule
     Method & Network & S & B & F \\
    \midrule
    \multirow{5}{*}{Grad} & ResNet34
     & 11.1 & 18.7 & \textbf{21.5} \\
    & \multirow{1}{*}{ResNet50}
     & 8.2 & 13.2 & \textbf{16.3} \\
    & \multirow{1}{*}{ResNet50 Robust}
     & 14.1 & 18.7 & \textbf{20.7}  \\
    & ResNet18 Camelyon16
     & 16.9 & 27.2 & \textbf{28.7}  \\     
    & \multirow{1}{*}{Total}
     & 12.6 & 19.5 & \textbf{21.8}  \\
     \midrule
    \multirow{1}{*}{DeepLift} & Total
     & 10.5 & 13.9 & \textbf{17.2} \\
     \midrule
    \multirow{1}{*}{IG} & Total
     & 8.6 & 12.2 & \textbf{15.7} \\
    \bottomrule
    \end{tabular}
    }
    \end{sc}
    \end{small}
    \end{center}
    \vskip -0.1in
\end{table}

\subsection{Total Variation}

In this section we report our results on the total variation experiments.
\cref{fig:TV_imagenet_layers} shows the total variation of the gradient throughout hidden layers of the ImageNet1K models (going deeper into the model from left to right).
It shows the result for the original model, and the three methods used to reduce the checkerboard artifacts (bilinear surrogate, backward hook, forward hook).
We can observe that all three methods reduce the total variation at every layer compared to using the original model.

\cref{tab:tv_aggregate} shows the percentage reduction in total variation of the saliency map compared to the original model.
On the right side of the table S stands for bilinear surrogate, B for backward hook and F for forward hook.
For the gradient we report both the total variation per model (averaged over all layers) as well as the total (averaged over all models).
Through all models and methods we see the largest reduction in TV for the ResNet18 Camelyon16 model, and the smallest reduction in the ResNet50 model.

\begin{figure*}[tb]
\begin{center}
\centerline{\includegraphics[width=\textwidth]{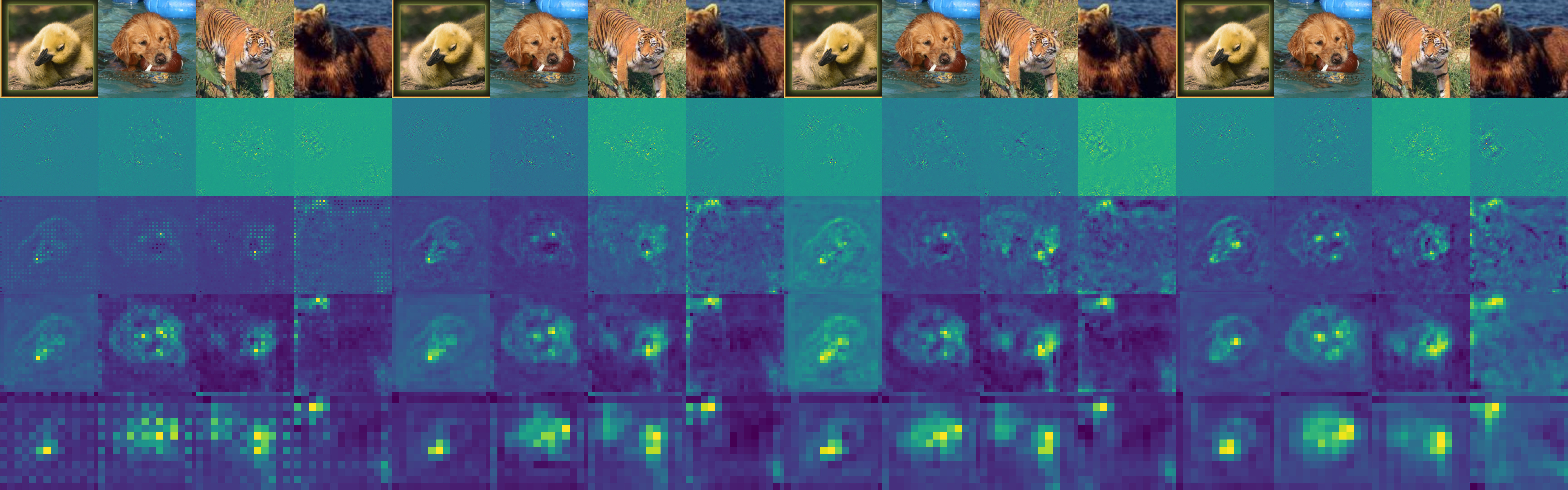}}
\caption{Saliency map from DeepLift from original model (first 4 images from the left), bilinear surrogate model (second 4 images from the left), backward hooked model (third 4 images) and forward hooked model (last 4 images). For the ResNet34. From top to bottom shows the original input image, saliency map for the input layer, Layer1, Layer2, Layer3.}
\label{fig:example_resnet50_robust}
\end{center}
\vskip -0.2in
\end{figure*}

\begin{figure}[tb]
\begin{center}
\centerline{\includegraphics[width=0.75\textwidth]{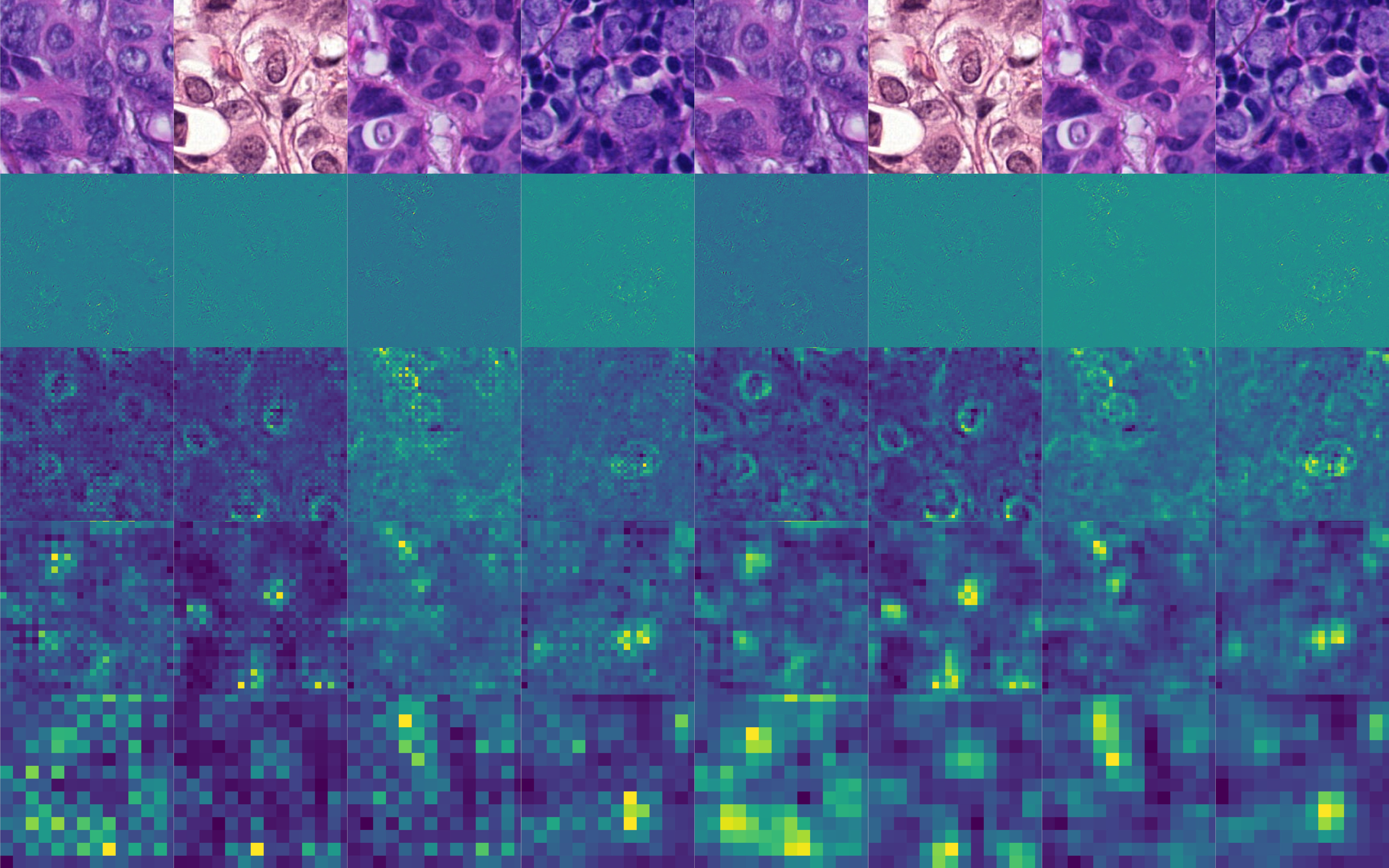}}
\caption{Saliency map from DeepLift from original (left 4 images), and bilinear surrogate model (right 4 images). For a ResNet18 on Camelyon16 data.}
\label{fig:example_resnet18_camelyon}
\end{center}
\vskip -0.2in
\end{figure}

\begin{figure}[tb]
\begin{center}
\centerline{\includegraphics[width=0.75\textwidth]{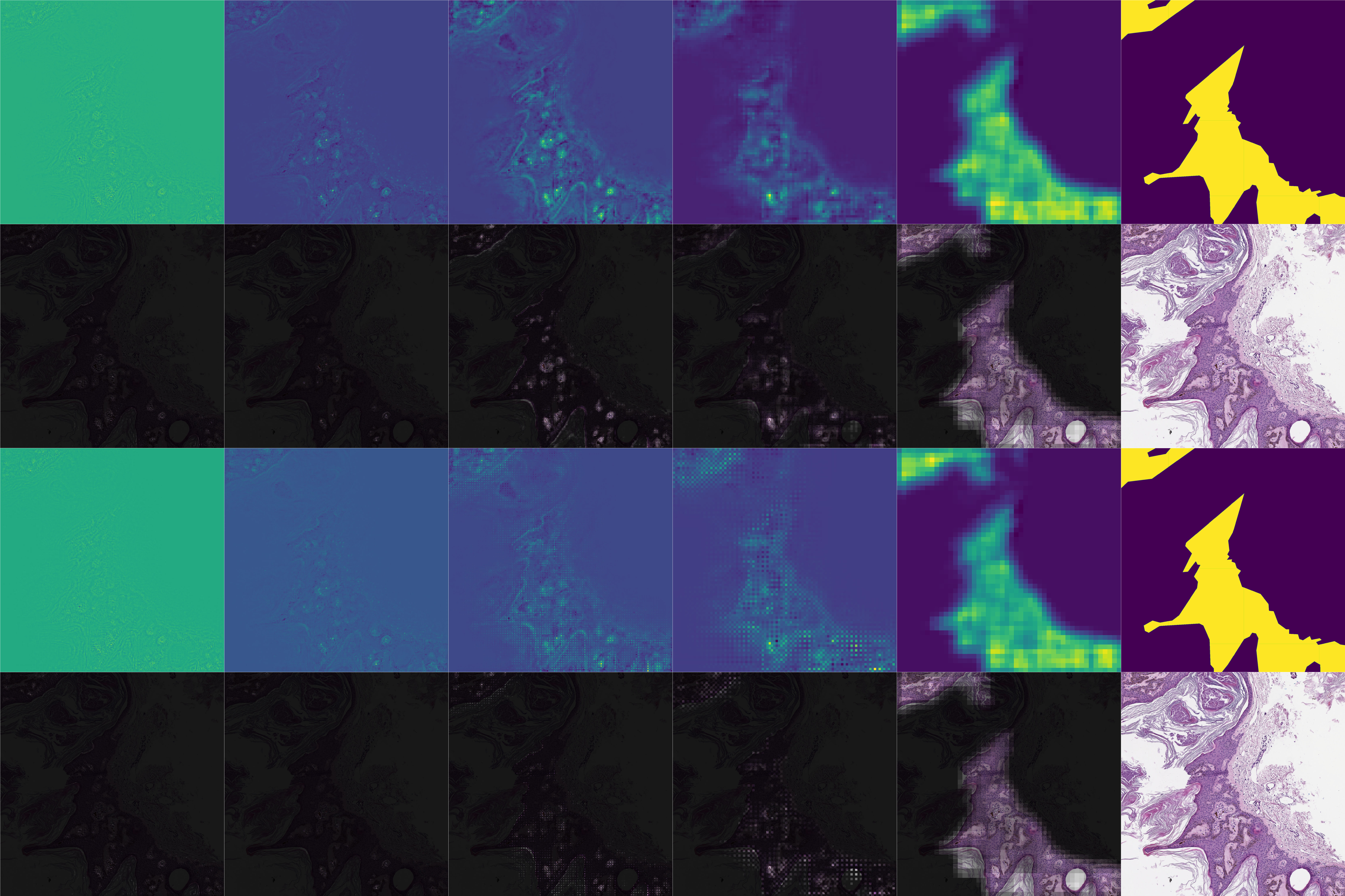}}
\caption{Saliency map from DeepLift for the ResNet34 on Digipath data. From left to right is: For the input layer, for Layer1, Layer2, Layer3, Layer4, ground truth tumor mask and input image. The top two rows are for the bilinear surrogate model, the bottom two rows are for the original model.}
\label{fig:example_digipath}
\end{center}
\vskip -0.2in
\end{figure}

\begin{figure}[tb]
\begin{center}
\centerline{\includegraphics[width=0.75\textwidth]{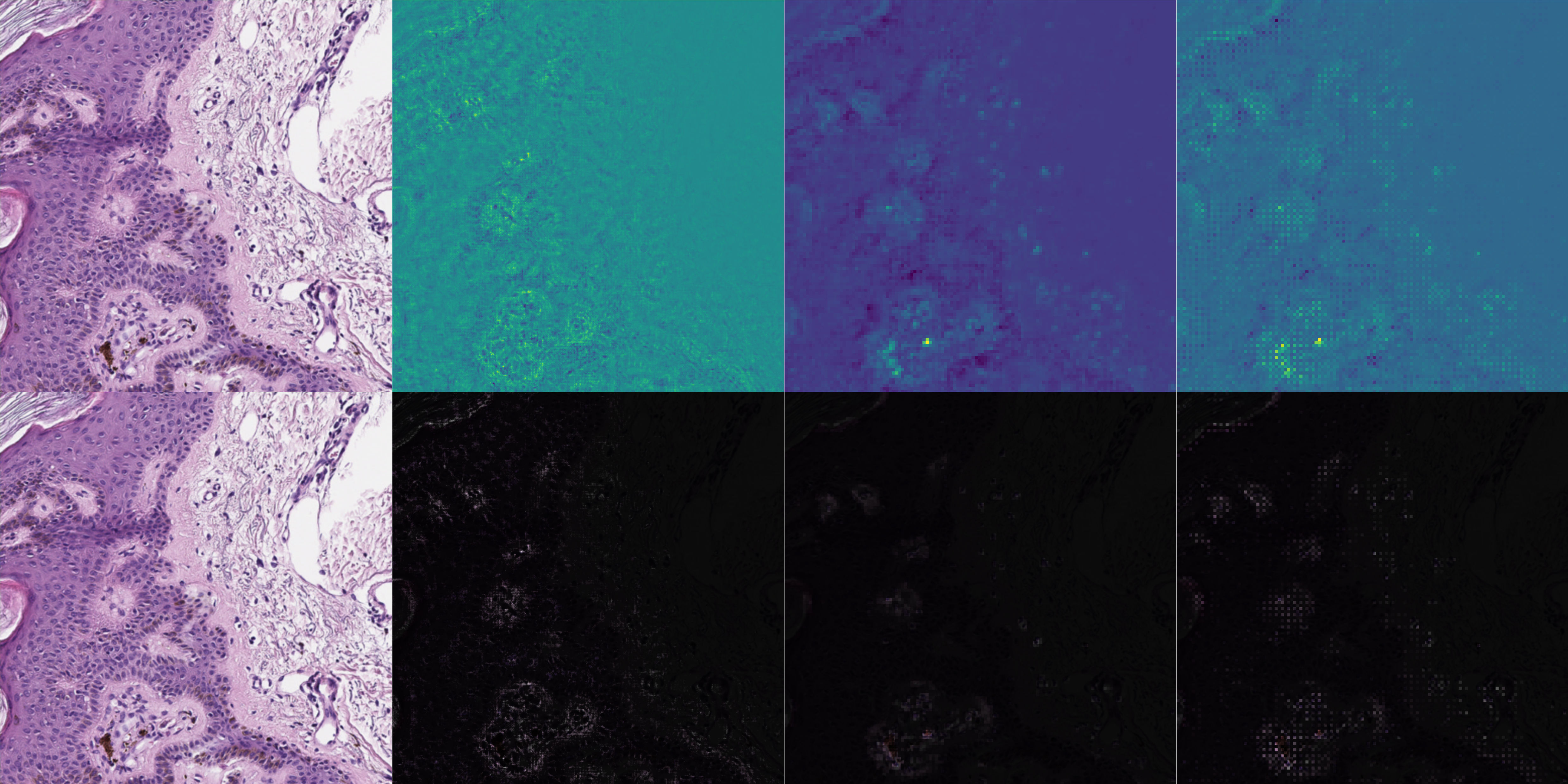}}
\caption{Saliency map from DeepLift on Digipath data, zoomed in on \cref{fig:example_digipath}. From left to right is: Input image, saliency map for the original model in the input layer, saliency map of the surrogate model for Layer1, saliency map of the original model for Layer1.}
\label{fig:example_digipath_zoomed}
\end{center}
\vskip -0.2in
\end{figure}

\subsection{Case Studies}

In this section we look at and discuss some example images.
The saliency maps have the spatial resolution of the layer they have been computed at, which is lower than the input resolution.
Therefore we upscale them to the size of the input.

An example for the usefulness of using saliency maps in hidden layers, is shown in \cref{fig:example_digipath} and \cref{fig:example_digipath_zoomed}.
\cref{fig:example_digipath} shows the saliency map for a single input image, for both the original model (bottom two rows) and the bilinear surrogate model (top two rows) for the input layer, Layer1, Layer2, Layer3, Layer4 (from left to right).
The top of the two rows is the saliency map, the bottom is overlaying it with the input image.
The saliency map for the input layer is highlighting edges and noise, and does not give much information about which structures the network uses to make its prediction.
Similarly, the saliency map for Layer4, which is the last convolutional layer (in the encoder of the model), is also not very helpful here, because it is just highlighting the whole tumor.
Since the model is a semantic segmentation network, this does not add more information than the prediction already gives.

The saliency map for the hidden layers for the bilinear surrogate model does add additional information though.
In Layer1, we can see individual cell nuclei being highlighted, telling us which kind of cells the model is using to support its prediction.
In Layer2, and Layer3 we can see structures formed by the cell nuclei.

\cref{fig:example_digipath_zoomed} only shows the center 512x512 pixels of the 1536x1536 images of \cref{fig:example_digipath} for the input layer and Layer1.
There we can see the highlighted cell nuclei more clearly.

\cref{fig:example_resnet50_robust} shows saliency maps for the ResNet34.
From left to right in pairs of 4 images it shows the saliency maps for the original model, the bilinear surrogate model, the backward hooked model and the forward hooked model.
From top to bottom it shows the original input, saliency map for the input, Layer1, Layer2 and Layer3.
The saliency map at the input is smooth in all cases.
But for the deeper layers there is a strong checkerboard noise pattern for the original model, whereas the maps for the other methods (bilinear surrogate, backward hooked model, forward hooked model) are smooth.

Something similar can be observed for \cref{fig:example_resnet18_camelyon}.
The figure shows saliency maps for the ResNet18 Camelyon16 for the original (left 4 images) and bilinear surrogate model (right 4 images) for Camelyon16 samples.
From top to bottom it shows the input image, saliency map for the input, for Layer1, Layer2, Layer3.
The model classified all 4 images as tumor.
The input layer is noisy and difficult to interpret.
In the deeper layers, the original model exhibits checkerboard noise in its saliency maps, whereas the saliency maps for the bilinear surrogate model are smoother and easier to interpret.
In Layer1 we can see cell nuclei being outlined, whereas deeper in Layer2 we can see the segmentations of whole cell nuclei.

\section{Limitations}
\label{limitations}

For the bilinear surrogate method to work, we need access to training data, in order to train the surrogate model.
Further, since all our methods reduce noise coming from convolutional downsampling, they are only applicable for models that utilize convolutional downsampling.
And since we are not computing the saliency map in the input but in a hidden layer and then overlay it with the input, we assume that the activations in hidden layers are localized (i.e. what the model extracts in a hidden layer and what we see in the input at those spots is roughly the same).

\section{Conclusion}
\label{conclusion}

We successfully investigated three methods to remove checkerboard noise of saliency maps in hidden layers of models that utilize convolutional downsampling.
The bilinear surrogate method closely matches the accuracy and predictions of the original model, while reducing the noise in the saliency map, measured by TV.
We observed a reduction in TV through all measured layers, with an average reduction of 10.5\% for DeepLift.
It performs slightly better under insertion metrics (2.14\%) compared to the original model, but a bit worse under deletion metrics (11.4\%).
The backward hook method also considerably reduces the TV and needs no training of a surrogate model, but it performs slightly worse under insertion and deletion metrics compared to the bilinear surrogate.
While the forward hook method has the lowest TV, and therefore smoothest saliency maps, it has considerably lower accuracy compared to the original model, its predictions differ rather strongly and it is not very faithful under insertion and deletion metrics.
Overall we recommend first trying the backward hook method, which removes the checkerboard noise and requires no additional training.
Then if the saliency map from hidden layers gives useful results, train a bilinear surrogate model.

%

\section{Acknowledgements}
R.H. is funded by the Deutsche Forschungsgemeinschaft (DFG, German Research Foundation) - Projectnumber 459360854 (DFG FOR 5347 Lifespan AI). M.S. and D.O.B. acknowledge the financial support by the German Federal Ministry of Economic Affairs and Climate Action (BMWK) and the European Social Fund (ESF) within the EXIST transfer of research project "aisencia" and the financial support by the Federal Ministry of Education and Research (BMBF) within the T!Raum-Initiative "\#MOIN! Modellregion Industriemathematik" in the sub-project "MUKIDerm".

\FloatBarrier

%
%
\bibliographystyle{splncs04}
\bibliography{main}

\section{Appendix}

\FloatBarrier

\begin{figure*}[tb]
    \centering
    \begin{tabular}{cccc}
        \begin{subfigure}{0.23\textwidth}
            \centering
            \includegraphics[width=\textwidth]{images/model_randomization/randomized_resnet34_layer1_del.png}
        \end{subfigure} &
        \begin{subfigure}{0.23\textwidth}
            \centering
            \includegraphics[width=\textwidth]{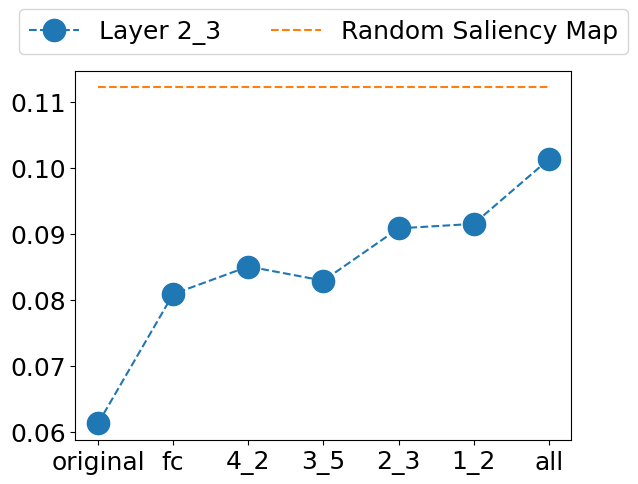}
        \end{subfigure} &
        \begin{subfigure}{0.23\textwidth}
            \centering
            \includegraphics[width=\textwidth]{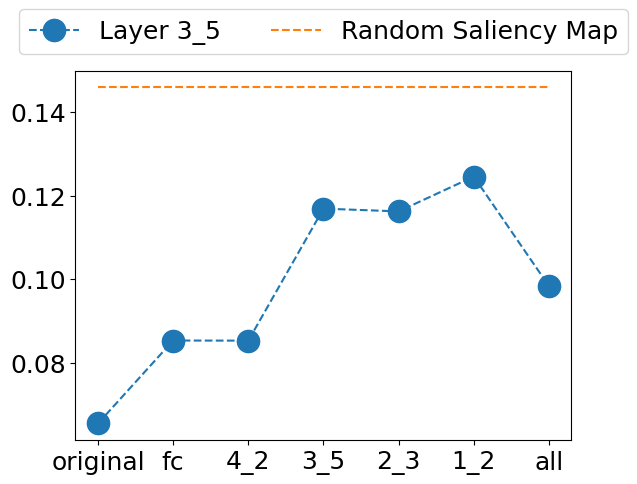}
        \end{subfigure} &
        \begin{subfigure}{0.23\textwidth}
            \centering
            \includegraphics[width=\textwidth]{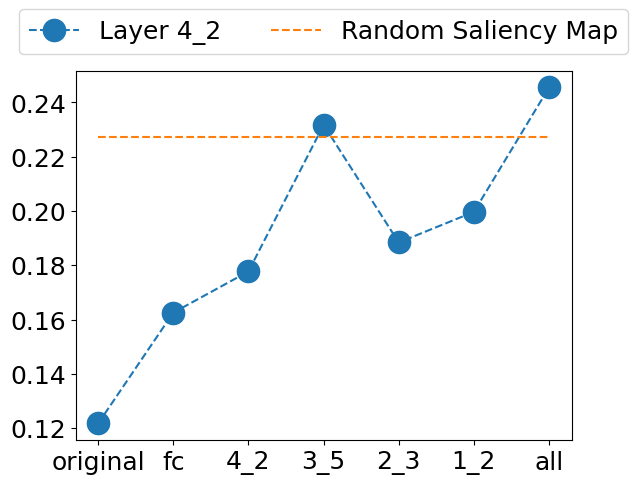}
        \end{subfigure} \\
        
        \begin{subfigure}{0.23\textwidth}
            \centering
            \includegraphics[width=\textwidth]{images/model_randomization/randomized_resnet34_layer1_ins.png}
        \end{subfigure} &
        \begin{subfigure}{0.23\textwidth}
            \centering
            \includegraphics[width=\textwidth]{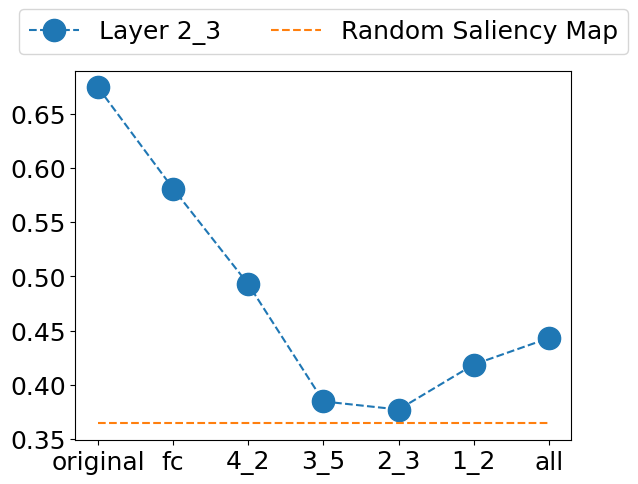}
        \end{subfigure} &
        \begin{subfigure}{0.23\textwidth}
            \centering
            \includegraphics[width=\textwidth]{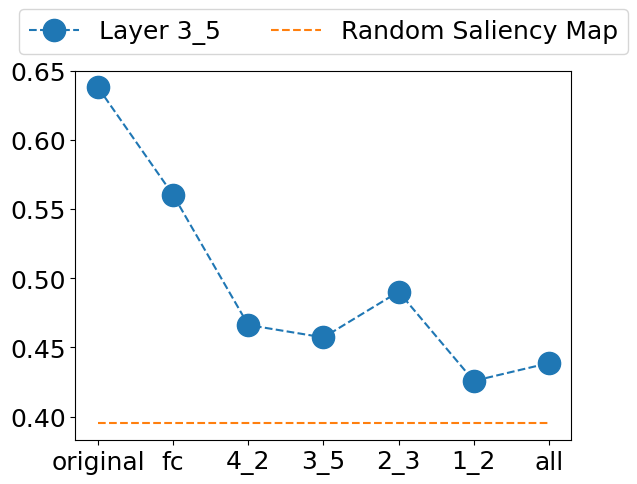}
        \end{subfigure} &
        \begin{subfigure}{0.23\textwidth}
            \centering
            \includegraphics[width=\textwidth]{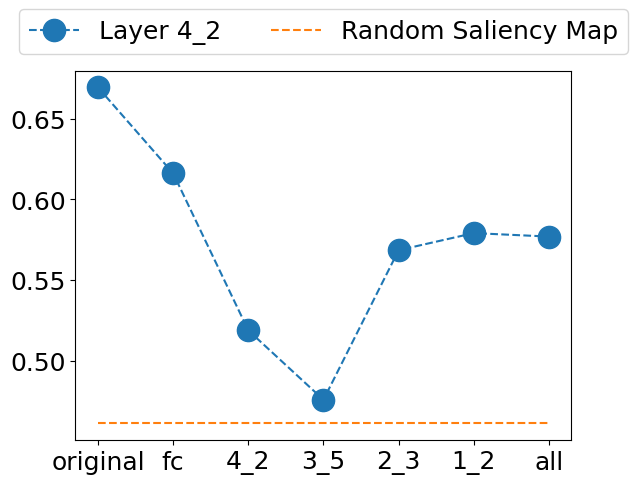}
        \end{subfigure}
    \end{tabular}
    \caption{Insertion and Deletion scores when randomizing the ResNet34, using DeepLift. Top row shows the deletion scores, bottom row shows the insertion scores}
    \label{fig:randomize_ResNet34}
\end{figure*}

\begin{figure*}[tb]
    \centering
    \begin{tabular}{cccc}
        \begin{subfigure}{0.23\textwidth}
            \centering
            \includegraphics[width=\textwidth]{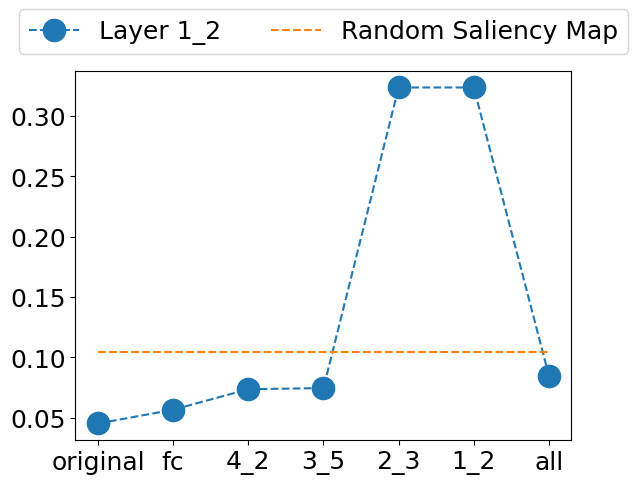}
        \end{subfigure} &
        \begin{subfigure}{0.23\textwidth}
            \centering
            \includegraphics[width=\textwidth]{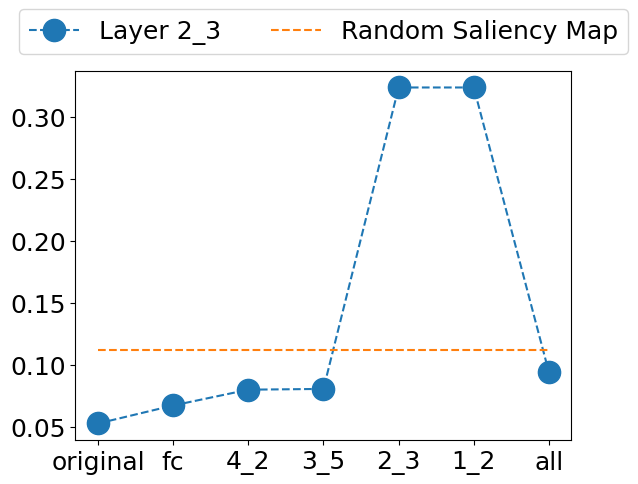}
        \end{subfigure} &
        \begin{subfigure}{0.23\textwidth}
            \centering
            \includegraphics[width=\textwidth]{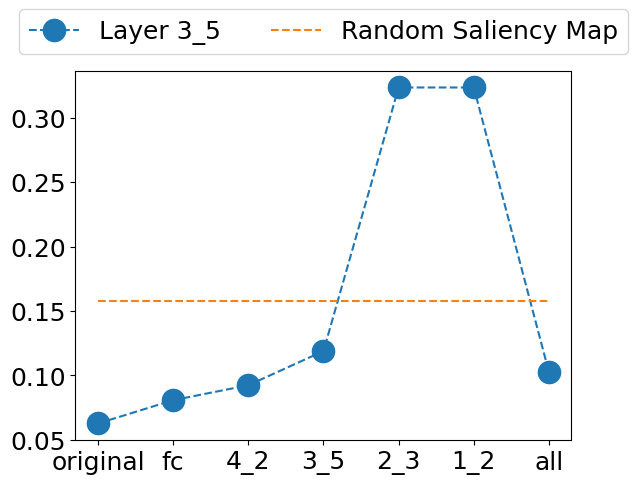}
        \end{subfigure} &
        \begin{subfigure}{0.23\textwidth}
            \centering
            \includegraphics[width=\textwidth]{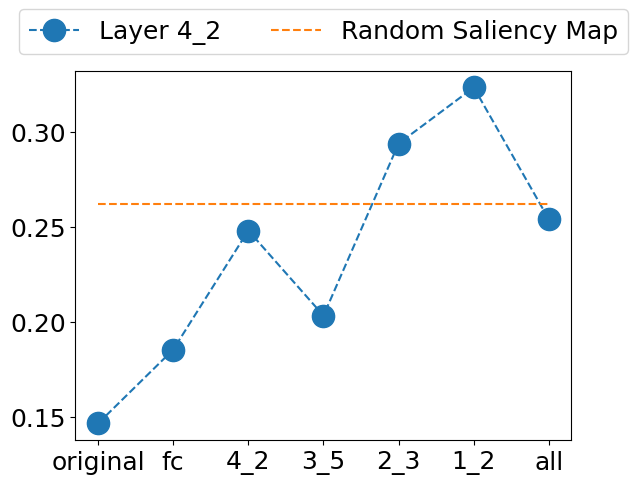}
        \end{subfigure} \\
        
        \begin{subfigure}{0.23\textwidth}
            \centering
            \includegraphics[width=\textwidth]{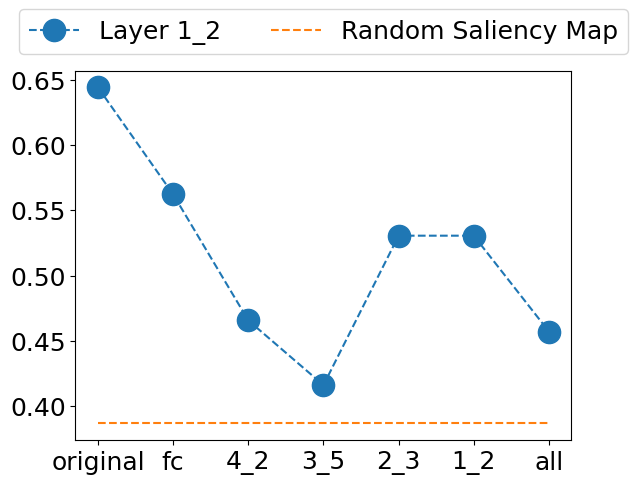}
        \end{subfigure} &
        \begin{subfigure}{0.23\textwidth}
            \centering
            \includegraphics[width=\textwidth]{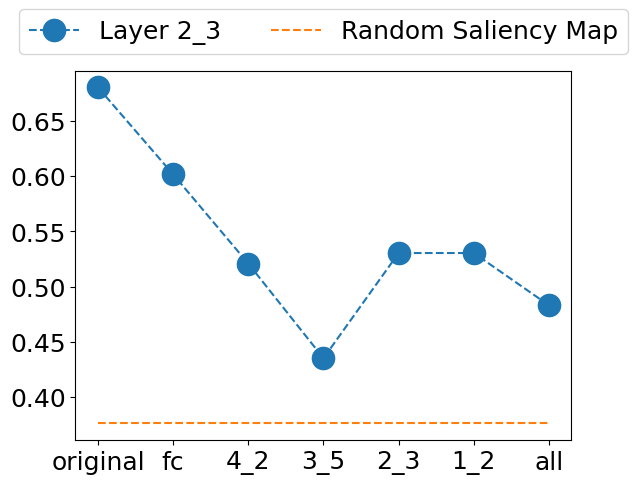}
        \end{subfigure} &
        \begin{subfigure}{0.23\textwidth}
            \centering
            \includegraphics[width=\textwidth]{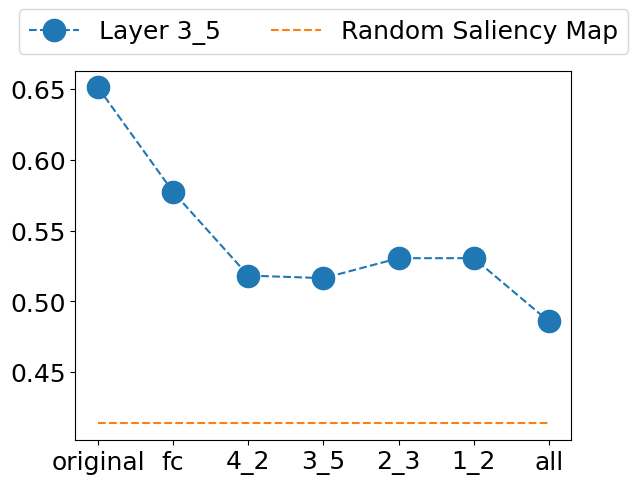}
        \end{subfigure} &
        \begin{subfigure}{0.23\textwidth}
            \centering
            \includegraphics[width=\textwidth]{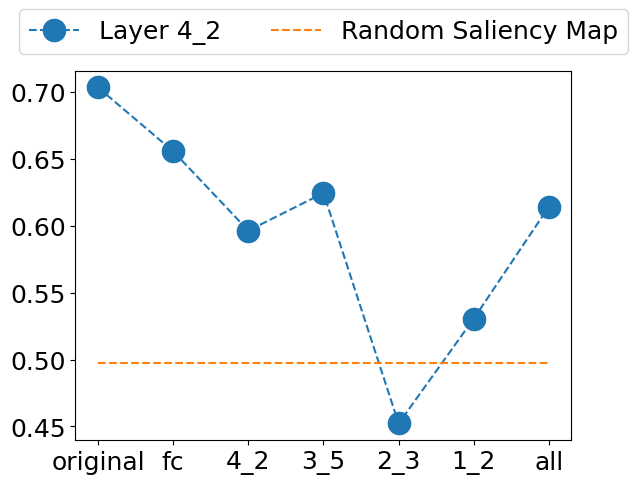}
        \end{subfigure}
    \end{tabular}
    \caption{Insertion and Deletion scores when randomizing the ResNet50, using DeepLift. Top row shows the deletion scores, bottom row shows the insertion scores}
    \label{fig:randomize_ResNet50}
\end{figure*}

\begin{figure*}[tb]
    \centering
    \begin{tabular}{cccc}
        \begin{subfigure}{0.23\textwidth}
            \centering
            \includegraphics[width=\textwidth]{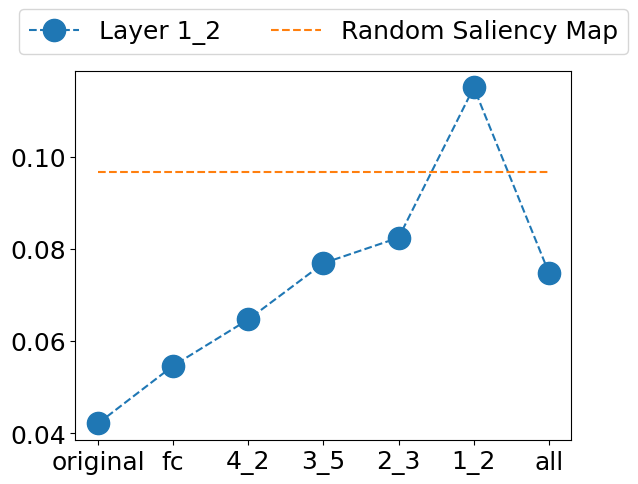}
        \end{subfigure} &
        \begin{subfigure}{0.23\textwidth}
            \centering
            \includegraphics[width=\textwidth]{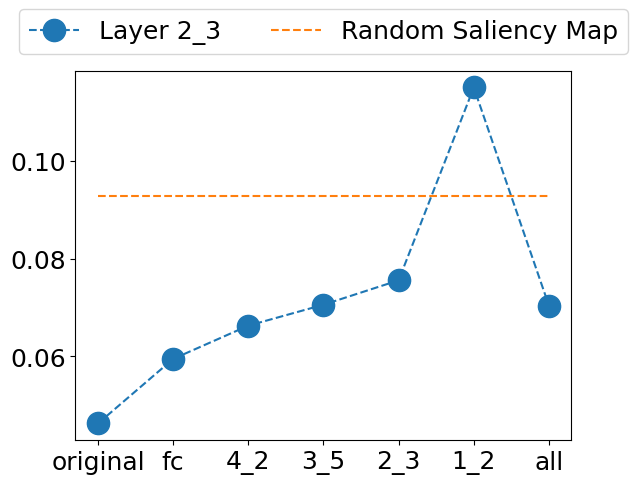}
        \end{subfigure} &
        \begin{subfigure}{0.23\textwidth}
            \centering
            \includegraphics[width=\textwidth]{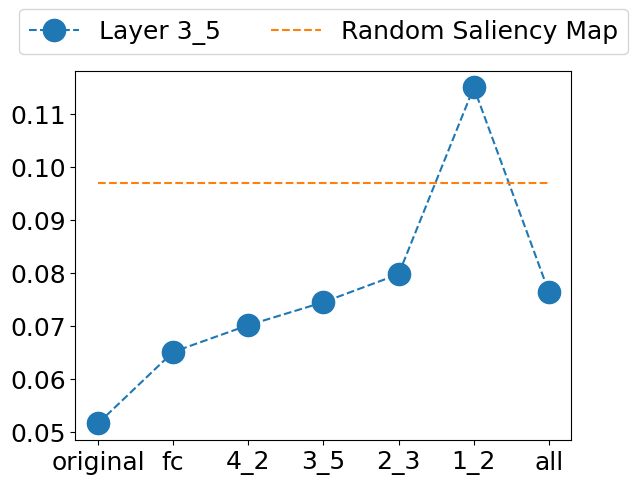}
        \end{subfigure} &
        \begin{subfigure}{0.23\textwidth}
            \centering
            \includegraphics[width=\textwidth]{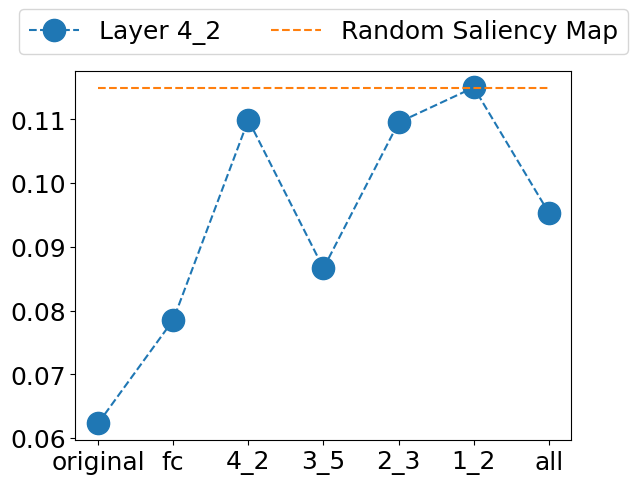}
        \end{subfigure} \\
        
        \begin{subfigure}{0.23\textwidth}
            \centering
            \includegraphics[width=\textwidth]{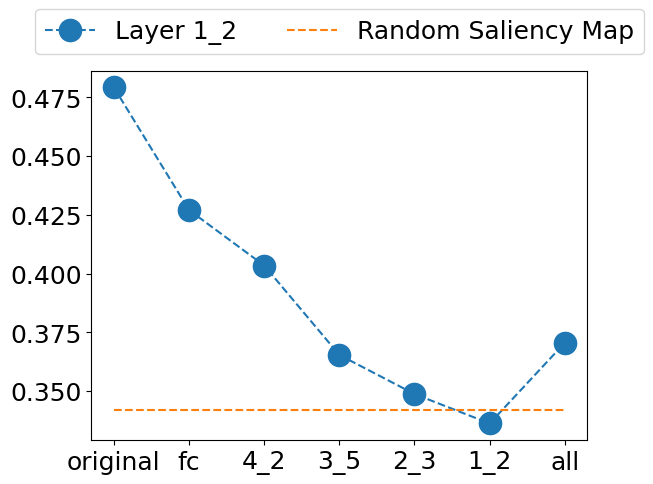}
        \end{subfigure} &
        \begin{subfigure}{0.23\textwidth}
            \centering
            \includegraphics[width=\textwidth]{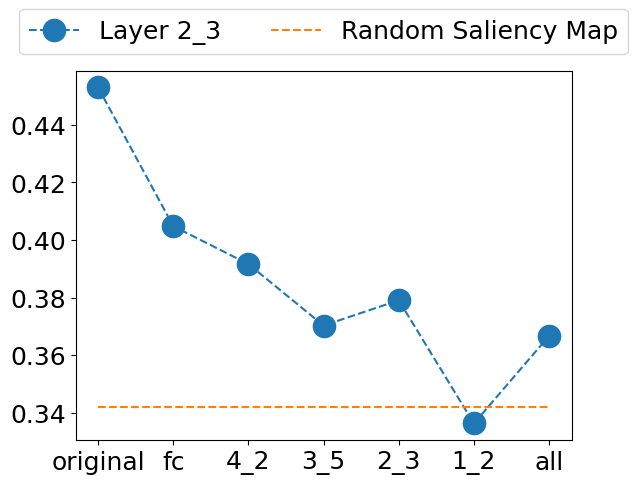}
        \end{subfigure} &
        \begin{subfigure}{0.23\textwidth}
            \centering
            \includegraphics[width=\textwidth]{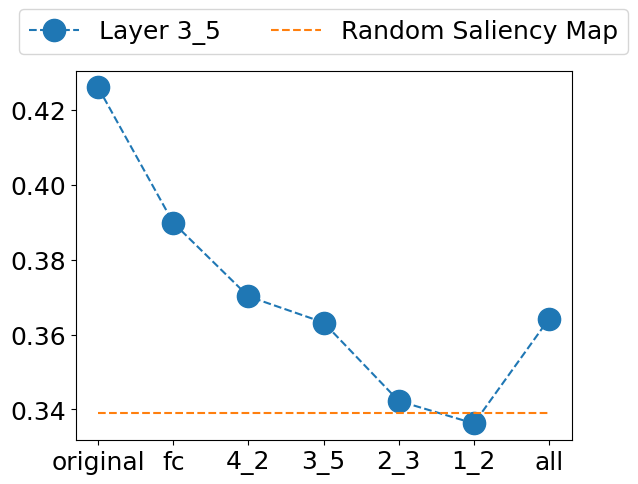}
        \end{subfigure} &
        \begin{subfigure}{0.23\textwidth}
            \centering
            \includegraphics[width=\textwidth]{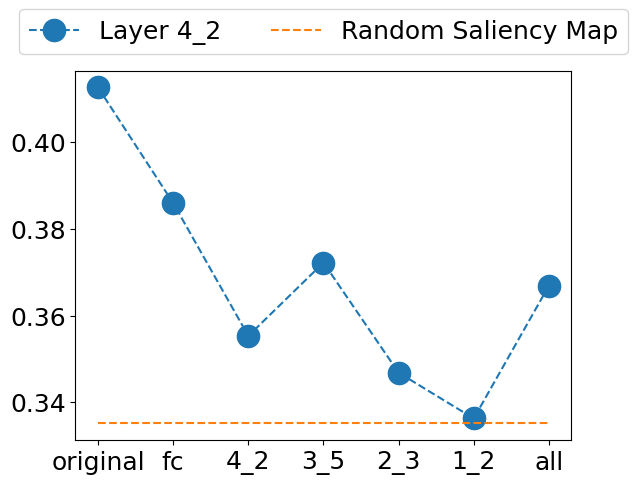}
        \end{subfigure}
    \end{tabular}
    \caption{Insertion and Deletion scores when randomizing the ResNet50 Robust, using DeepLift. Top row shows the deletion scores, bottom row shows the insertion scores}
    \label{fig:randomize_ResNet50_Robust}
\end{figure*}

In this appendix we show additional figures from the experiments.
\cref{tab:tv_aggregate_appendix} includes the total variation reduction for the individual models for DeepLift and IG (in the paper Table 5 only included the total result for those two methods).
\cref{fig:randomize_ResNet34} shows the insertion and deletion score when randomizing layers in the ResNet34, similar for \cref{fig:randomize_ResNet50} and \cref{fig:randomize_ResNet50_Robust} which show the results for the ResNet50 respectively the robust ResNet50.

\begin{figure*}[tb]
    \centering
    \begin{tabular}{cc}
        \begin{subfigure}{0.4\textwidth}
            \centering
            \includegraphics[width=\textwidth]{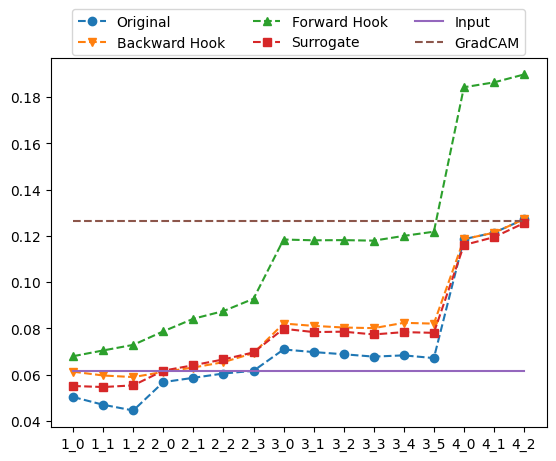}
            \caption*{ResNet34 del}
        \end{subfigure} &
        \begin{subfigure}{0.4\textwidth}
            \centering
            \includegraphics[width=\textwidth]{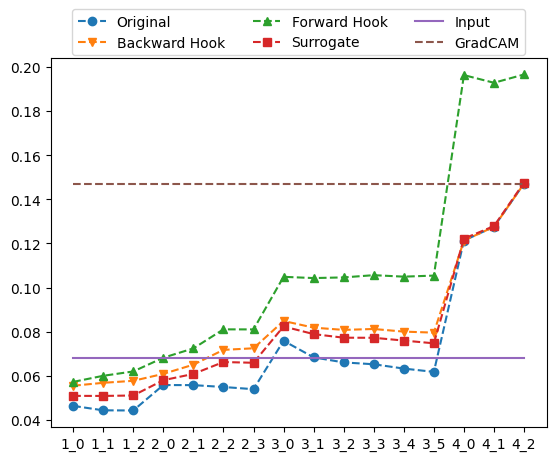}
            \caption*{ResNet50 del}
        \end{subfigure} \\
        
        \begin{subfigure}{0.4\textwidth}
            \centering
            \includegraphics[width=\textwidth]{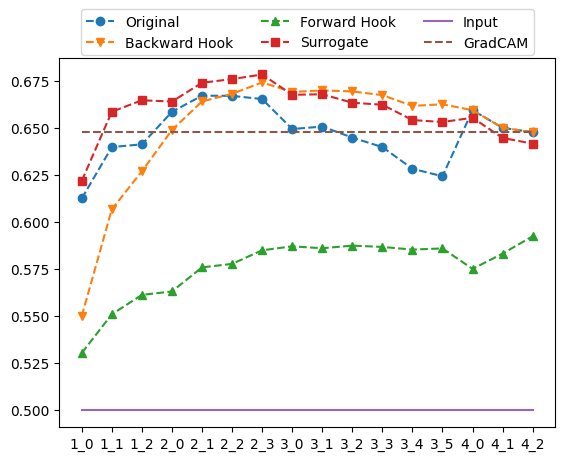}
            \caption*{ResNet34 ins}
        \end{subfigure} &
        \begin{subfigure}{0.4\textwidth}
            \centering
            \includegraphics[width=\textwidth]{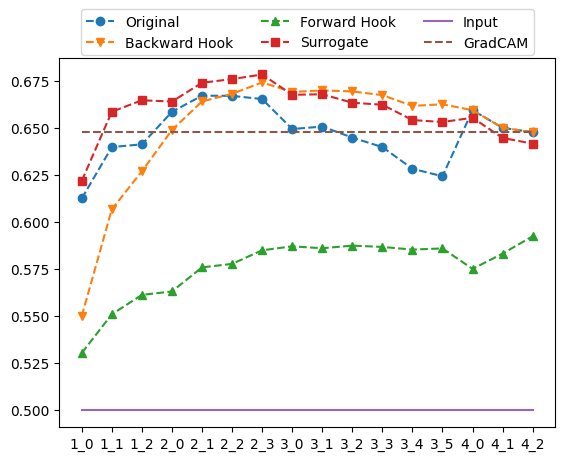}
            \caption*{ResNet50 ins}
        \end{subfigure}
    \end{tabular}
    \caption{Insertion and Deletion scores using DeepLift for the ResNet34 and ResNet50}
    \label{fig:DeepLift_ResNet34_ResNet50}
\end{figure*}

\begin{figure*}[tb]
    \centering
    \begin{tabular}{cc}
        \begin{subfigure}{0.4\textwidth}
            \centering
            \includegraphics[width=\textwidth]{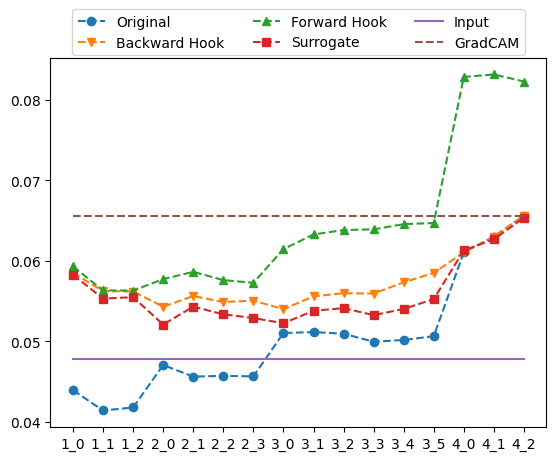}
            \caption*{ResNet50 Robust del}
        \end{subfigure} &
        \begin{subfigure}{0.4\textwidth}
            \centering
            \includegraphics[width=\textwidth]{images/experiments/metrics/resnet18_del.png}
            \caption*{ResNet18 Camelyon16 del}
        \end{subfigure} \\
        
        \begin{subfigure}{0.4\textwidth}
            \centering
            \includegraphics[width=\textwidth]{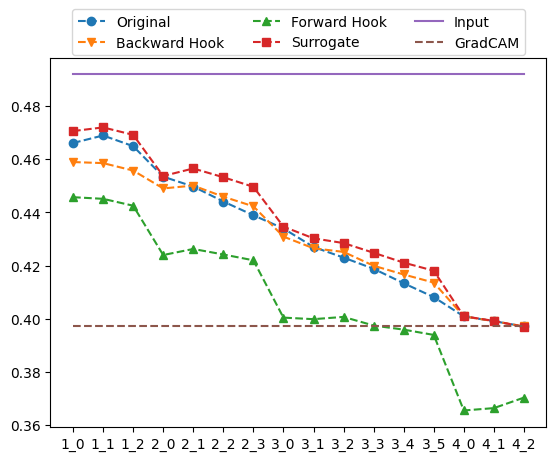}
            \caption*{ResNet50 Robust ins}
        \end{subfigure} &
        \begin{subfigure}{0.4\textwidth}
            \centering
            \includegraphics[width=\textwidth]{images/experiments/metrics/resnet18_ins.png}
            \caption*{ResNet18 Camelyon16 ins}
        \end{subfigure}
    \end{tabular}
    \caption{Insertion and Deletion scores using DeepLift for the ResNet50 Robust and the ResNet18 Camelyon16}
    \label{fig:DeepLift_ResNet50_Robust_ResNet18}
\end{figure*}

\begin{figure*}[tb]
    \centering
    \begin{tabular}{cc}
        \begin{subfigure}{0.4\textwidth}
            \centering
            \includegraphics[width=\textwidth]{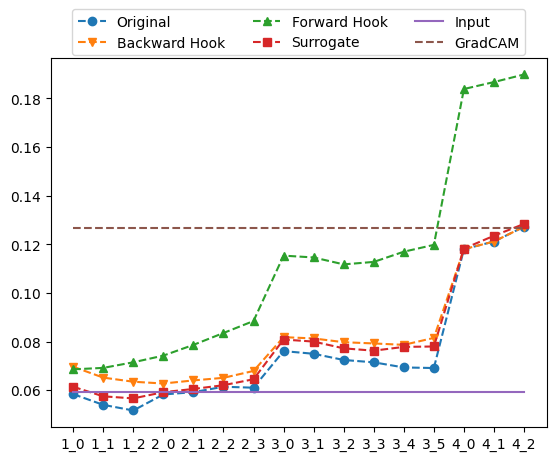}
            \caption*{ResNet34 del}
        \end{subfigure} &
        \begin{subfigure}{0.4\textwidth}
            \centering
            \includegraphics[width=\textwidth]{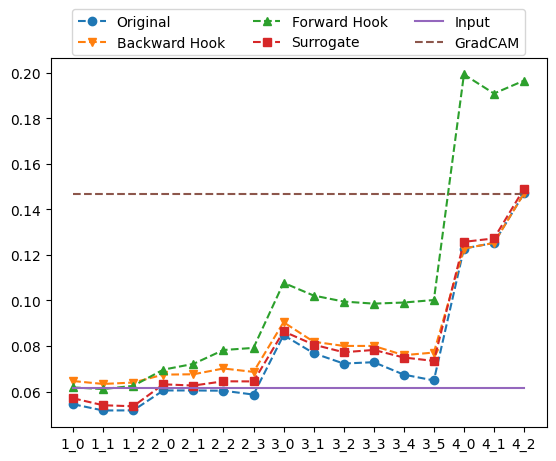}
            \caption*{ResNet50 del}
        \end{subfigure} \\
        
        \begin{subfigure}{0.4\textwidth}
            \centering
            \includegraphics[width=\textwidth]{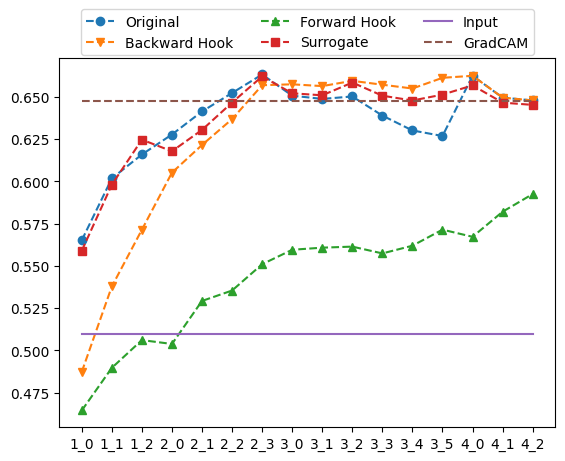}
            \caption*{ResNet34 ins}
        \end{subfigure} &
        \begin{subfigure}{0.4\textwidth}
            \centering
            \includegraphics[width=\textwidth]{images/experiments/metrics/resnet34_IG_ins.png}
            \caption*{ResNet50 ins}
        \end{subfigure}
    \end{tabular}
    \caption{Insertion and Deletion scores using Integrated Gradients for the ResNet34 and ResNet50}
    \label{fig:IG_ResNet34_ResNet50}
\end{figure*}

\begin{figure*}[tb]
    \centering
    \begin{tabular}{cc}
        \begin{subfigure}{0.4\textwidth}
            \centering
            \includegraphics[width=\textwidth]{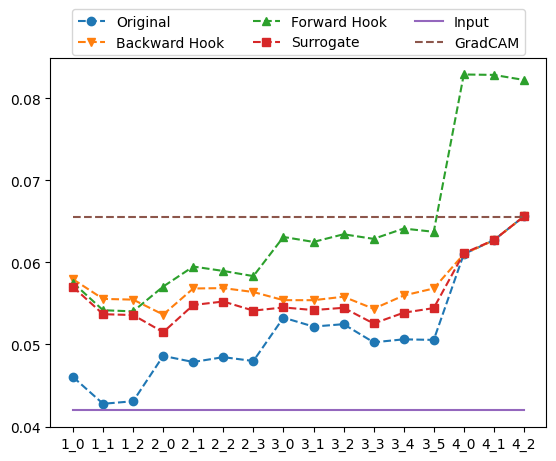}
            \caption*{ResNet50 Robust del}
        \end{subfigure} &
        \begin{subfigure}{0.4\textwidth}
            \centering
            \includegraphics[width=\textwidth]{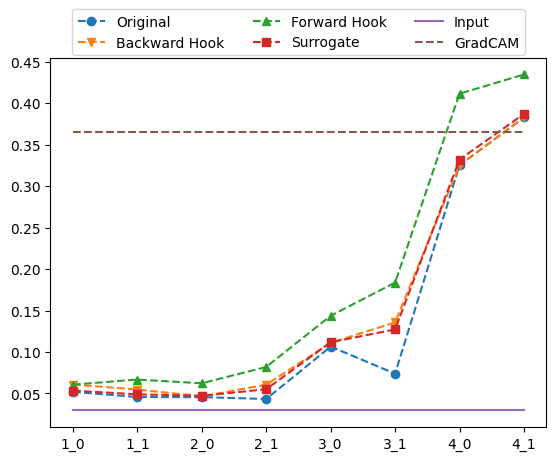}
            \caption*{ResNet18 Camelyon16 del}
        \end{subfigure} \\
        
        \begin{subfigure}{0.4\textwidth}
            \centering
            \includegraphics[width=\textwidth]{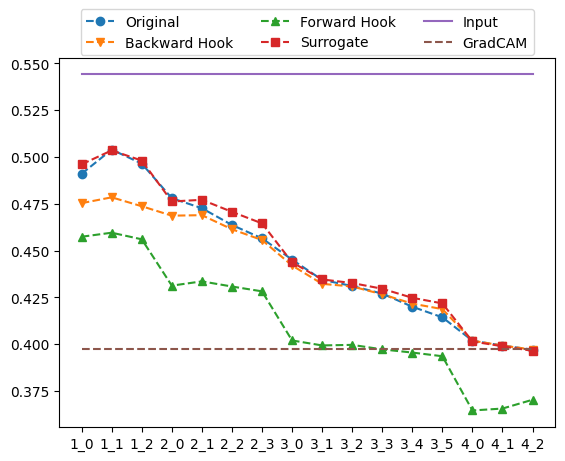}
            \caption*{ResNet50 Robust ins}
        \end{subfigure} &
        \begin{subfigure}{0.4\textwidth}
            \centering
            \includegraphics[width=\textwidth]{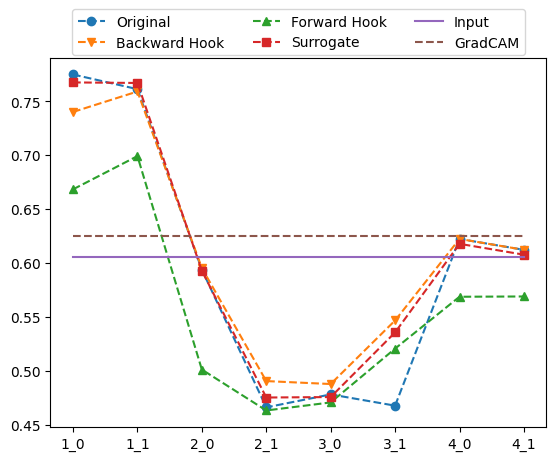}
            \caption*{ResNet18 Camelyon16 ins}
        \end{subfigure}
    \end{tabular}
    \caption{Insertion and Deletion scores using Integrated Gradients for the ResNet50 Robust and the ResNet18 Camelyon16}
    \label{fig:IG_ResNet50_Robust_ResNet18}
\end{figure*}

\begin{figure*}[tb]
    \centering
    \begin{tabular}{cc}
        \begin{subfigure}{0.4\textwidth}
            \centering
            \includegraphics[width=\textwidth]{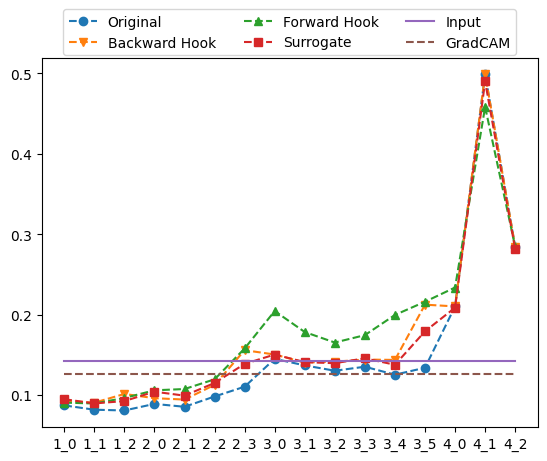}
            \caption*{ResNet34 del}
        \end{subfigure} &
        \begin{subfigure}{0.4\textwidth}
            \centering
            \includegraphics[width=\textwidth]{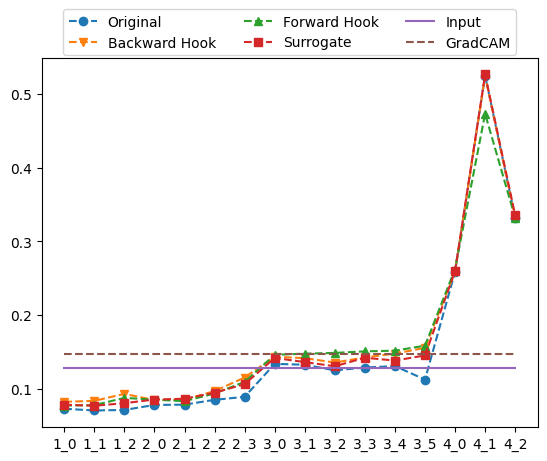}
            \caption*{ResNet50 del}
        \end{subfigure} \\
        
        \begin{subfigure}{0.4\textwidth}
            \centering
            \includegraphics[width=\textwidth]{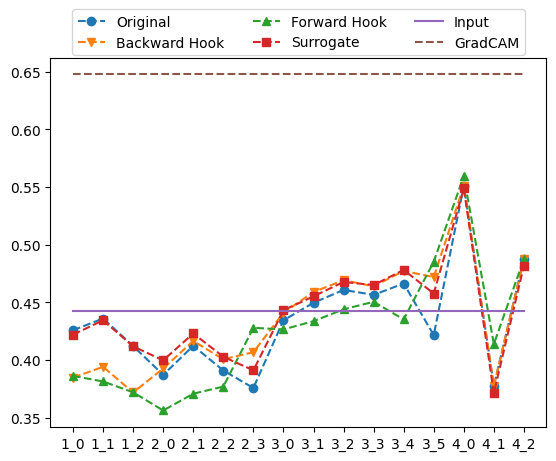}
            \caption*{ResNet34 ins}
        \end{subfigure} &
        \begin{subfigure}{0.4\textwidth}
            \centering
            \includegraphics[width=\textwidth]{images/experiments/metrics/resnet34_Grad_ins.png}
            \caption*{ResNet50 ins}
        \end{subfigure}
    \end{tabular}
    \caption{Insertion and Deletion scores using the gradient for the ResNet34 and ResNet50}
    \label{fig:Grad_ResNet34_ResNet50}
\end{figure*}

\begin{figure*}[tb]
    \centering
    \begin{tabular}{cc}
        \begin{subfigure}{0.4\textwidth}
            \centering
            \includegraphics[width=\textwidth]{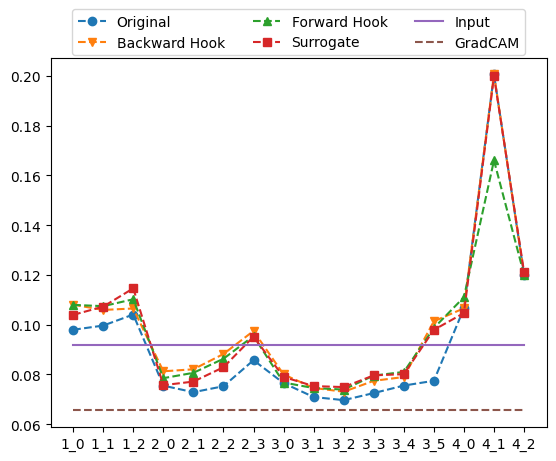}
            \caption*{ResNet50 Robust del}
        \end{subfigure} &
        \begin{subfigure}{0.4\textwidth}
            \centering
            \includegraphics[width=\textwidth]{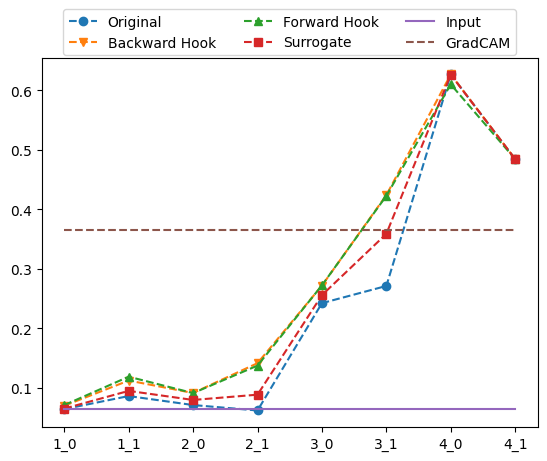}
            \caption*{ResNet18 Camelyon16 del}
        \end{subfigure} \\
        
        \begin{subfigure}{0.4\textwidth}
            \centering
            \includegraphics[width=\textwidth]{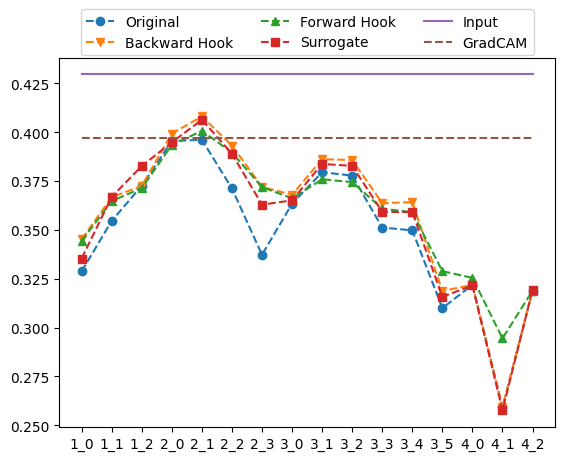}
            \caption*{ResNet50 Robust ins}
        \end{subfigure} &
        \begin{subfigure}{0.4\textwidth}
            \centering
            \includegraphics[width=\textwidth]{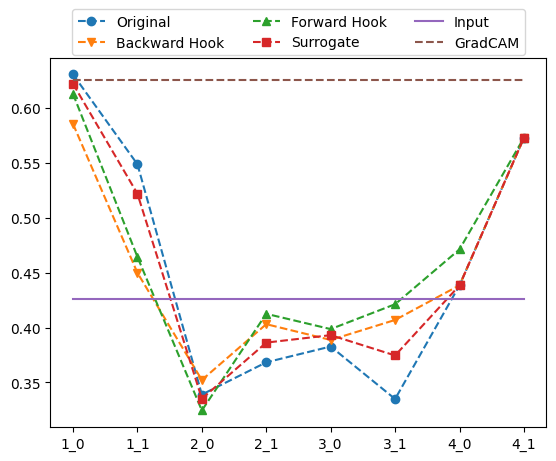}
            \caption*{ResNet18 Camelyon16 ins}
        \end{subfigure}
    \end{tabular}
    \caption{Insertion and Deletion scores using the gradient for the ResNet50 Robust and the ResNet18 Camelyon16}
    \label{fig:Grad_ResNet50_Robust_ResNet18}
\end{figure*}

\cref{fig:DeepLift_ResNet34_ResNet50} to \cref{fig:Grad_ResNet50_Robust_ResNet18} show the results from the insertion and deletion metrics for DeepLift, Integrated Gradients and simply using the gradient for the three ImageNet1K models and the ResNet18 trained on Camelyon16 (the paper only showed the results for DeepLift).

\begin{figure*}[htb]
    \centering
    \begin{subfigure}{0.3\textwidth}
        \centering
        \includegraphics[width=\textwidth]{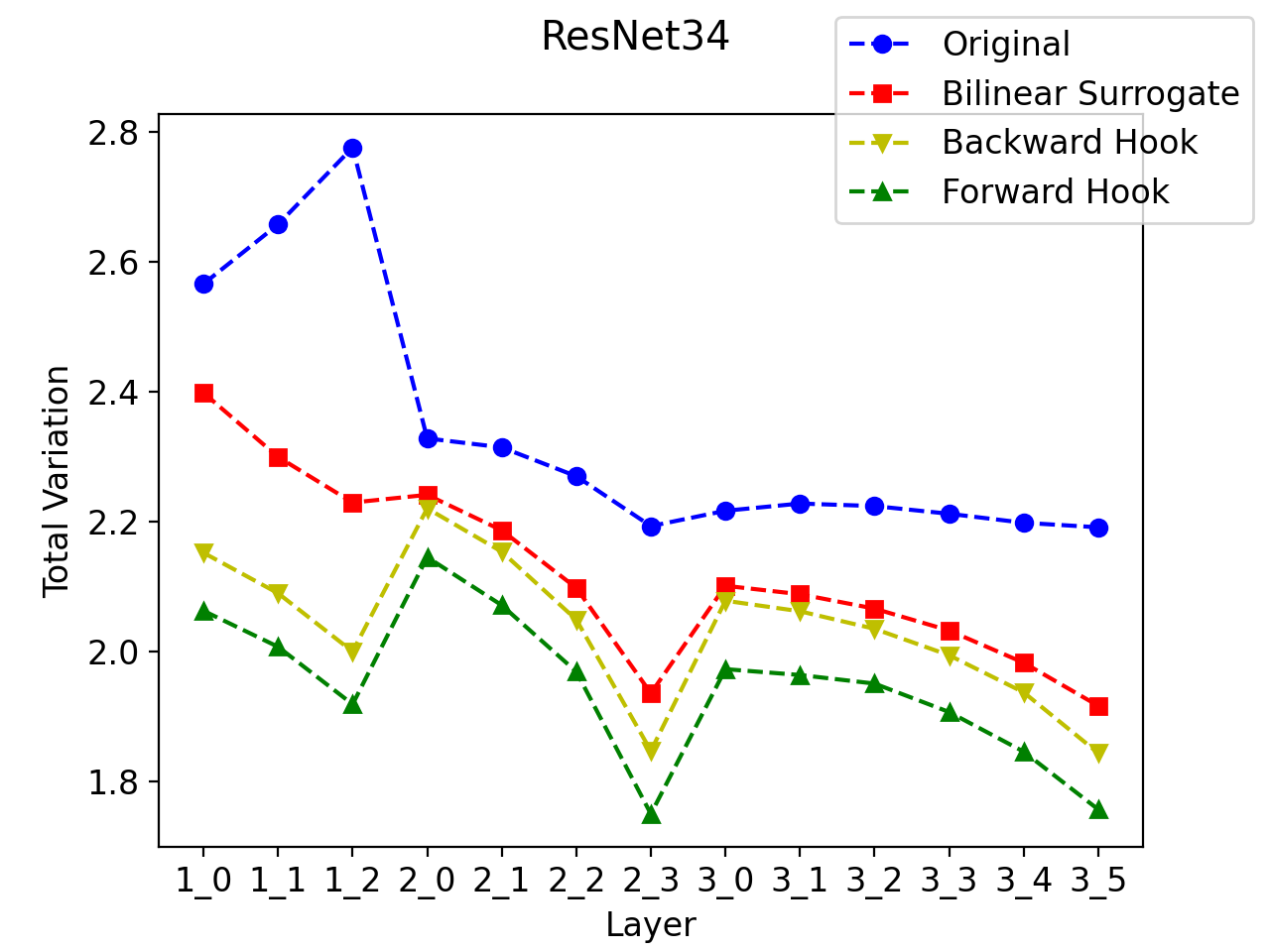}
    \end{subfigure}
    \begin{subfigure}{0.3\textwidth}
        \centering
        \includegraphics[width=\textwidth]{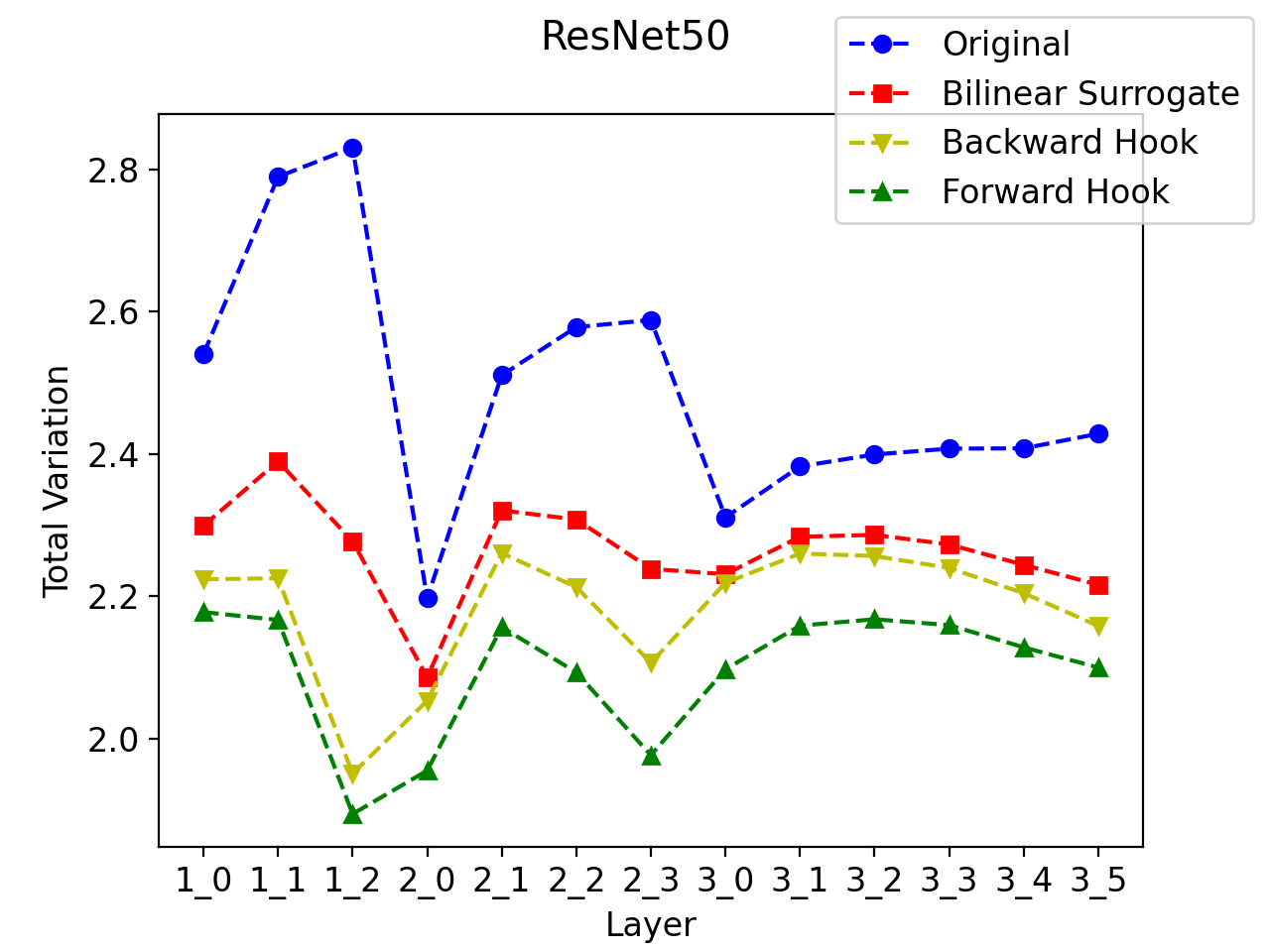}
    \end{subfigure}
    \begin{subfigure}{0.3\textwidth}
        \centering
        \includegraphics[width=\textwidth]{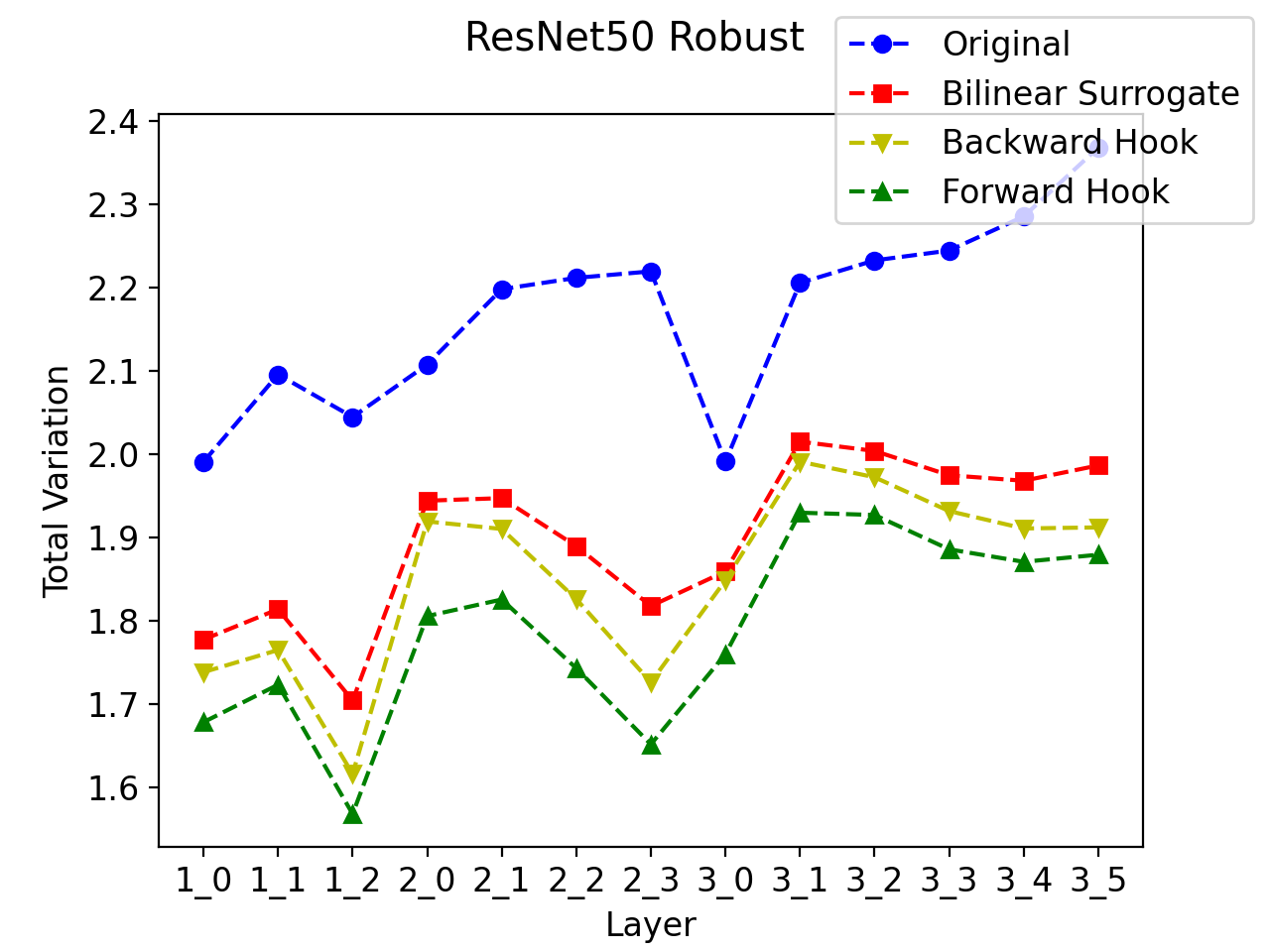}
    \end{subfigure}
    \caption{Total variation of the DeepLift saliency map method.}
    \label{fig:appendix_TV_imagenet_layers_DeeepLift}
    \label{fig:TV_DeepLift}
\end{figure*}

\begin{figure*}[htb]
    \centering
    \begin{subfigure}{0.3\textwidth}
        \centering
        \includegraphics[width=\textwidth]{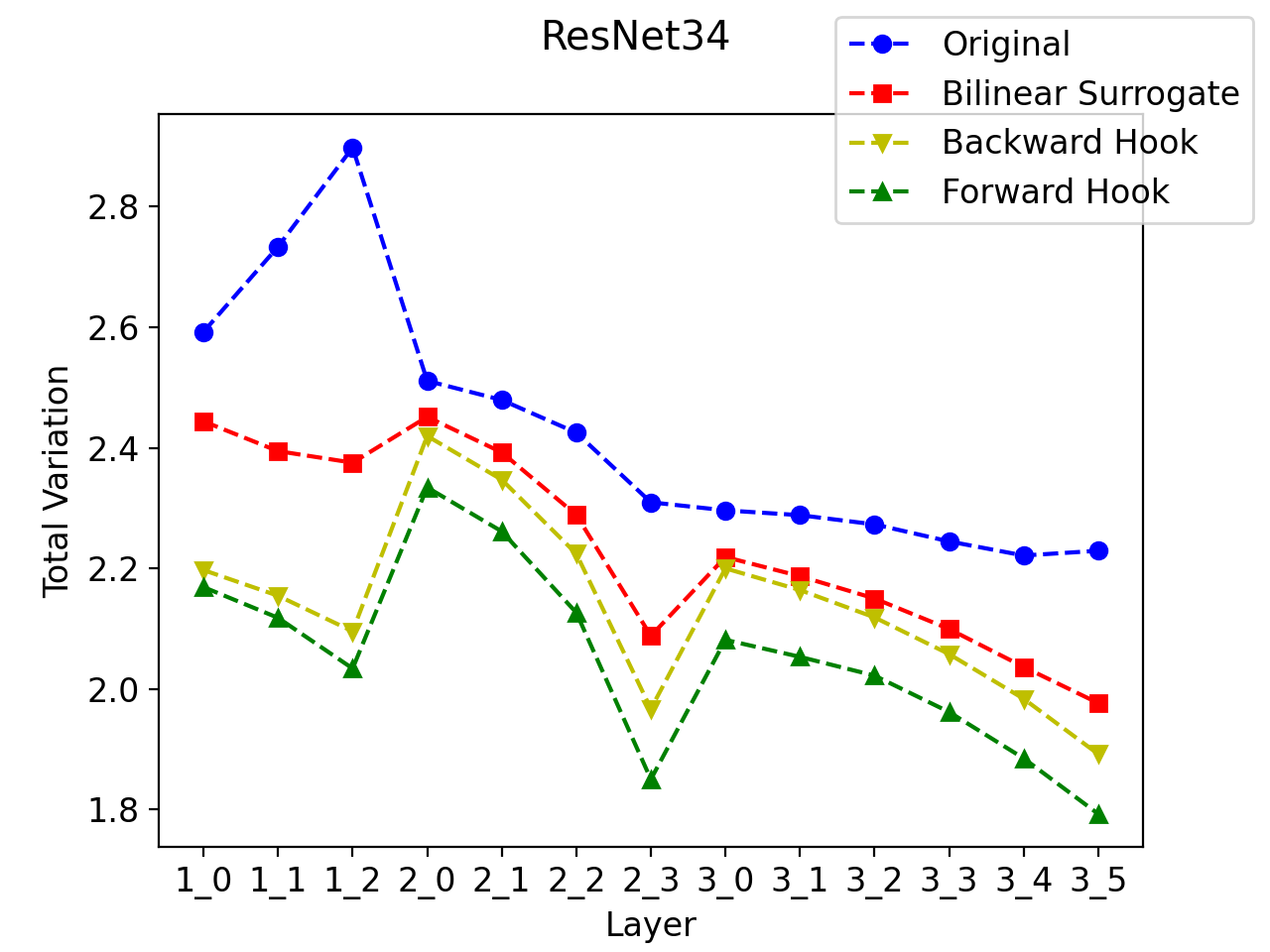}
    \end{subfigure}
    \begin{subfigure}{0.3\textwidth}
        \centering
        \includegraphics[width=\textwidth]{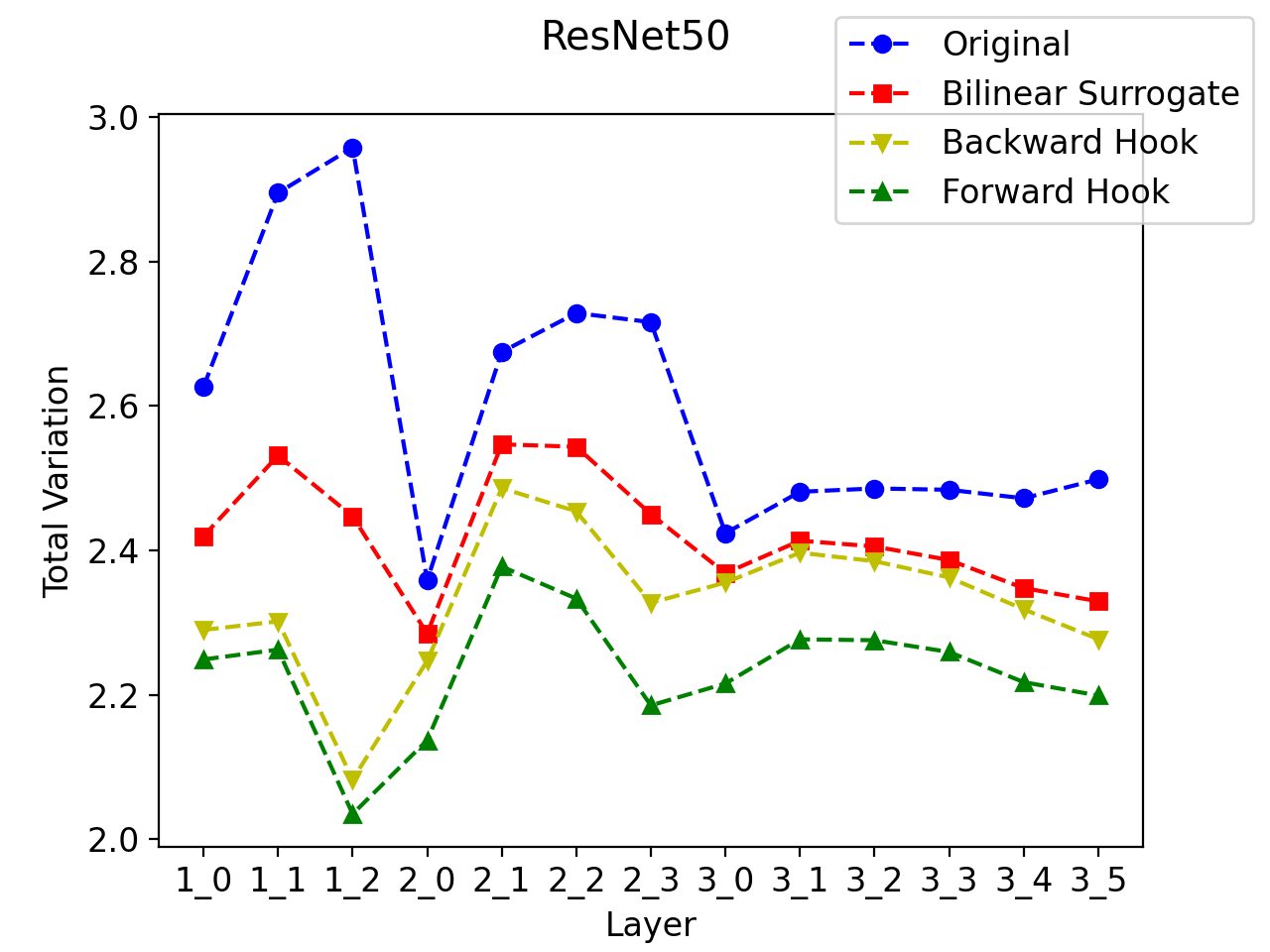}
    \end{subfigure}
    \begin{subfigure}{0.3\textwidth}
        \centering
        \includegraphics[width=\textwidth]{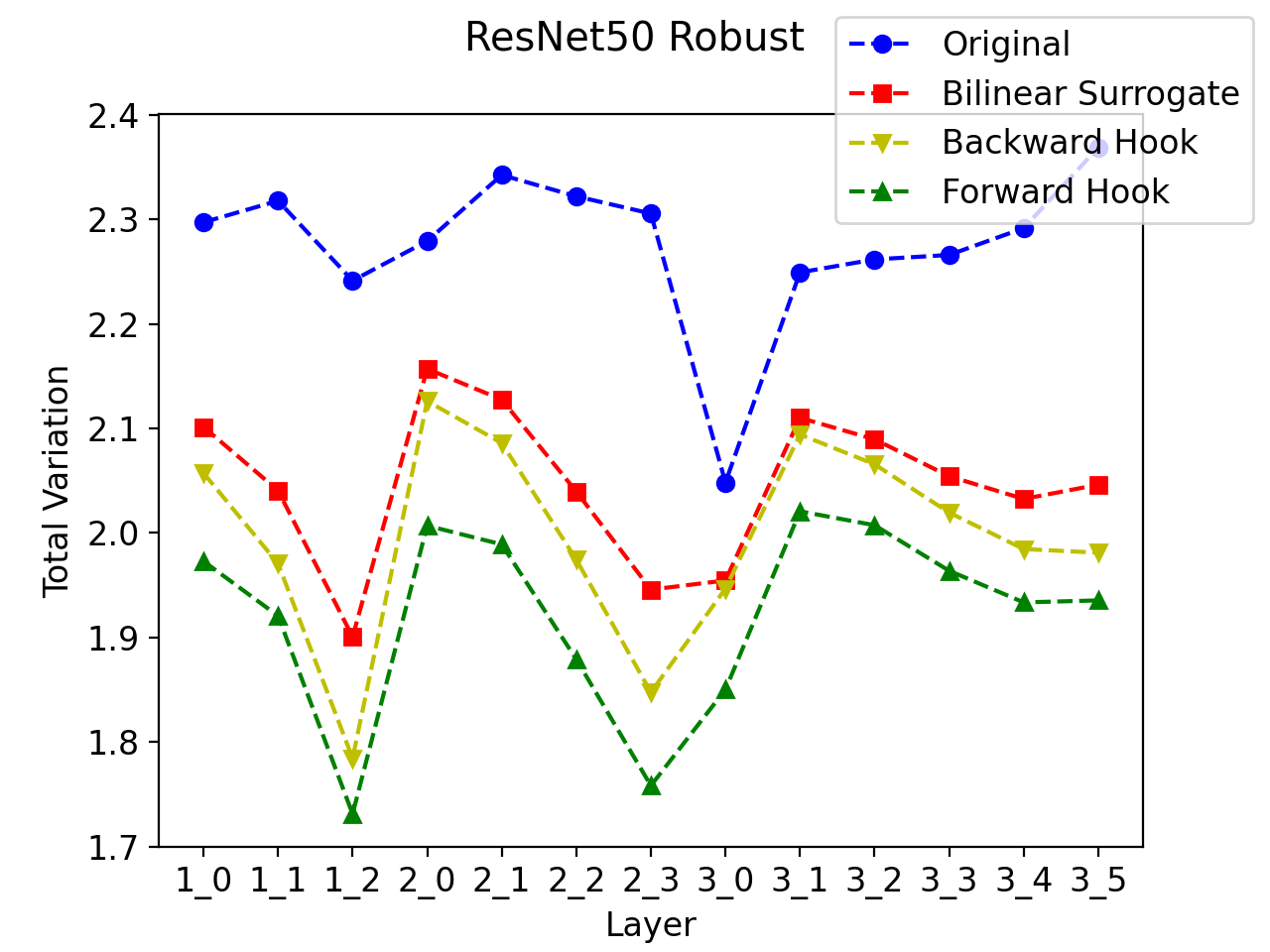}
    \end{subfigure}
    \caption{Total variation of the Integrated Gradients saliency map method.}
    \label{fig:appendix_TV_imagenet_layers_IG}
    \label{fig:TV_IG}
\end{figure*}

\cref{fig:appendix_TV_imagenet_layers_DeeepLift} and \cref{fig:appendix_TV_imagenet_layers_IG} show the total variation of the saliency map when using DeepLift respectively Integrated Gradients as saliency method (the paper only included the result for using the gradient).

\begin{table}[tb]
    \caption{Percentage reduction in total variation of the saliency map compared to the original model}
    \label{tab:tv_aggregate_appendix}
    \begin{center}
    \begin{small}
    \begin{sc}
    \vskip -0.2in
    \resizebox{.5\textwidth}{!}{
    \begin{tabular}{l|l|cccc}  
    \toprule
     Method & Network & S & B & F \\
    \midrule
    \multirow{5}{*}{Grad} & ResNet34
     & 11.1 & 18.7 & \textbf{21.5} \\
    & \multirow{1}{*}{ResNet50}
     & 8.2 & 13.2 & \textbf{16.3} \\
    & \multirow{1}{*}{ResNet50 Robust}
     & 14.1 & 18.7 & \textbf{20.7}  \\
    & ResNet18 Camelyon16
     & 16.9 & 27.2 & \textbf{28.7}  \\     
    & \multirow{1}{*}{Total}
     & 12.6 & 19.5 & \textbf{21.8}  \\
     \midrule
    \multirow{5}{*}{DeepLift} & ResNet34
     & 9.0 & 12.5 & \textbf{16.3} \\
    & \multirow{1}{*}{ResNet50}
     & 8.7 & 11.9 & \textbf{15.5} \\
    & \multirow{1}{*}{ResNet50 Robust}
     & 12.3 & 14.6 & \textbf{17.5}  \\
    & ResNet18 Camelyon16
     & 11.8 & 16.5 & \textbf{19.5}  \\  
     & Total
     & 10.5 & 13.9 & \textbf{17.2} \\
     \midrule
    \multirow{5}{*}{IG} & ResNet34
     & 7.4 & 11.3 & \textbf{15.0} \\
    & \multirow{1}{*}{ResNet50}
     & 6.6 & 9.9 & \textbf{13.7} \\
    & \multirow{1}{*}{ResNet50 Robust}
     & 10.0 & 12.3 & \textbf{15.6}  \\
    & ResNet18 Camelyon16
     & 10.4 & 15.1 & \textbf{18.3}  \\  
     & Total
     & 8.6 & 12.2 & \textbf{15.7} \\
    \bottomrule
    \end{tabular}
    }
    \end{sc}
    \end{small}
    \end{center}
    \vskip -0.1in
\end{table}

\end{document}